\documentclass[10pt,journal]{IEEEtran}

\usepackage[T1]{fontenc}
\usepackage{cite}
\usepackage{amsmath,amssymb,amsfonts}
\usepackage{graphicx}
\usepackage{booktabs}
\usepackage{multirow}
\usepackage{xcolor}
\usepackage{enumitem}
\usepackage[hidelinks]{hyperref}
\usepackage{microtype}
\usepackage{tikz}
\usepackage{array}
\usepackage{multirow}
\usepackage{authblk}

\title{World Models for Robotic Manipulation: A Survey}
\author[1]{Fangyuan Wang\textsuperscript{\dag}}
\author[2,3]{Ziyuan Wang\textsuperscript{\dag}}
\author[4]{Guorui Pei}
\author[5]{Mengshi Zhang}
\author[6]{Canxi Liang}
\author[3]{Jun Hu}
\author[7]{Zhongxuan Li}
\author[1]{Jinsong Wu}
\author[1]{Ning Han}
\author[8]{Zeqing Zhang}
\author[9]{Jiaming Qi}
\author[10]{Hongmin Wu}
\author[3]{Shiyao Zhang}
\author[11]{Pai Zheng}
\author[7]{Jia Pan}
\author[1]{David Navarro-Alarcon}
\author[12]{Sichao Liu\textsuperscript{*}}
\author[3]{Peng Zhou\textsuperscript{*}}

\affil[1]{Department of Mechanical Engineering, The Hong Kong Polytechnic University, Kowloon, Hong Kong SAR, China}
\affil[2]{Department of Mechanical Engineering and Automation, Harbin Institute of Technology, Shenzhen, China}
\affil[3]{School of Advanced Engineering, Great Bay University, Dongguan, Guangdong, China}
\affil[4]{College of Robotics Science and Engineering, Taiyuan University of Technology, Taiyuan, China}
\affil[5]{School of Data Science, City University of Hong Kong (Dongguan), Dongguan, Guangdong, China}
\affil[6]{Department of Mechatronic Engineering, Guangdong Polytechnic Normal University, Guangdong, China}
\affil[7]{School of Computing and Data Science, The University of Hong Kong, Hong Kong SAR, China}
\affil[8]{School of Electrical and Electronic Engineering, Nanyang Technological University, Singapore}
\affil[9]{College of Mechanical and Electrical Engineering, Northeast Forestry University, Harbin, China}
\affil[10]{Greater Bay Area National Center of Technology Innovation, Guangzhou, China}
\affil[11]{Department of Industrial and Systems Engineering, The Hong Kong Polytechnic University, Kowloon, Hong Kong SAR}
\affil[12]{Department of Production Engineering, KTH Royal Institute of Technology, Stockholm, Sweden}

\affil[ ]{\textit{\textsuperscript{\dag} Co-first author \textsuperscript{*} Corresponding author}}

\begin{document}

\maketitle

\begin{abstract}
Robotic manipulation depends on the ability to anticipate how actions reshape objects, contacts, and scene geometry before execution. Learned world models provide this capability by predicting task-relevant future evolution under robot intervention, yet the term now spans latent dynamics models, action-conditioned video generators, three- and four-dimensional scene predictors, physics-informed simulators, and predictive modules inside vision-language-action systems. This breadth has fragmented the literature and obscured the design choices that matter for manipulation. We survey world models for robotic manipulation through three questions: what future representation is predicted, how prediction is connected to action, and when prediction is used in the robot-learning pipeline. We operationally define a world model as an action-conditioned predictive system and distinguish it from perception modules, inverse models, policies, rewards, and value functions. We then organize existing work into five representation families, develop a functional taxonomy that separates integrated prediction-action models from explicit predictive planners, and characterize infrastructure roles including synthetic experience generation, candidate filtering, search-based evaluation, learned environments, and outcome verification. We further map these roles across pretraining, post-training, and inference adaptation, review 34 manipulation datasets, and synthesize evaluation protocols for predictive fidelity, task performance, and simulator reliability. This survey shows that world models are evolving from task-specific dynamics predictors into predictive infrastructure for robot learning, while exposing open challenges in contact modeling, hallucination control, action alignment, and benchmarking under closed-loop use.
\end{abstract}


\section{Introduction}
\IEEEPARstart{R}{obotic} manipulation is a problem of decision making under physical interaction. A robot must act through contact, occlusion, partial observability, and irreversible state changes, often with only limited opportunities for trial and error. Acting only from the current observation is therefore insufficient. The robot must anticipate how objects, contacts, and scene geometry will evolve under its own actions. Classical work in motor control and model-based reinforcement learning addressed this need through internal forward models that estimate the sensory and physical consequences of motor commands~\cite{jordan1991internal,wolpert1995internal,sutton1991dyna,deisenroth2011lowcost}. The same idea underlies the modern notion of a \emph{world model}: a learned predictive system that estimates task-relevant future evolution and supplies that prediction to a policy, planner, simulator, or learning algorithm~\cite{ha2018worldmodels,planet2018,dreamer2019,dreamerv3}.

The scope of world modeling has expanded rapidly as robot learning has moved from compact task-specific controllers to large-scale, language-conditioned, generalist systems. Beyond the latent dynamics models of the Dreamer family, predictive systems now appear as action-conditioned video generators, learned simulators for policy post-training, three- and four-dimensional scene predictors, physics-informed dynamics models, and predictive components inside vision-language-action systems~\cite{seer,worldvla,world4rl,worldgym,genie_envisioner,tesseract,pointworld}. This expansion has made the term \emph{world model} increasingly ambiguous. In one community, it may refer to a compact transition model used for planning. In another, it may refer to a video generator that imagines possible futures. In a third, it may denote a learned environment, a reward-producing simulator, or a reasoning module embedded within a VLA policy. The shared idea is prediction, but the predicted representation, action interface, and stage of use often differ substantially.

This ambiguity is especially costly for manipulation. Predictive fidelity and action utility do not always coincide. A model may generate visually plausible futures while failing to preserve contact, object permanence, force closure, or action feasibility. Conversely, a compact latent model may support efficient control while hiding the physical structure needed for inspection, transfer, or safety-critical deployment. Different research communities resolve this tension in different ways. Model-based reinforcement learning emphasizes compact latent dynamics and sample-efficient planning. Video prediction emphasizes temporal coherence and visual realism. Geometric and physics-informed models prioritize spatial structure and physical plausibility. VLA systems entangle prediction with language-conditioned reasoning and action generation. A manipulation-centered survey must therefore separate three questions that are often conflated: what future representation is predicted, how that prediction affects action, and when the predictive model is used within the learning pipeline.

Recent surveys provide valuable context, but none fully resolves this manipulation-specific design space. General world-model surveys emphasize video generation and autonomous driving while treating manipulation only peripherally~\cite{zhu2024sora}. Surveys of embodied intelligence discuss world models alongside physical simulators as enabling technologies rather than as a unified predictive design space~\cite{long2025embodied}. Surveys of three- and four-dimensional modeling focus on geometric representations and do not cover the broader model-based reinforcement learning and imitation learning traditions~\cite{kong2025worldsurvey}. The most closely related manipulation survey either avoids committing to a strict definition and organizes methods by perception, prediction, and control~\cite{zhang2025stepwm}, or restricts attention to VLA agents and excludes much of the model-based reinforcement learning and imitation learning literature in which the modern world-model concept originated~\cite{tan_vla_survey}. Adjacent surveys of VLM-based VLAs and embodied VLA models treat world modeling as one auxiliary direction rather than as the central object of analysis~\cite{shao2025vlmvla,ma2024vlasurvey}. As a result, the field lacks a survey that connects reinforcement learning, imitation learning, video generation, geometry, physics, and VLA systems through a common manipulation-centered account of predictive modeling.

This survey addresses that gap by organizing world models for robotic manipulation around three complementary axes. First, we ask what the model represents. We compare image and video prediction, learned latent dynamics, motion fields, scene flow, three- and four-dimensional scene structure, and physics-informed dynamics. Second, we ask how prediction is connected to action. We distinguish integrated prediction-action models, in which prediction is fused with action generation, from explicit predictive planners that expose subgoals, trajectories, waypoints, latent plans, or hierarchical plans to a separate executor. Third, we ask when the model is used. We analyze world models as predictive priors during pretraining, as data engines, learned simulators, reward sources, and verifiers during post-training, and as online reasoning or adaptation mechanisms during inference. Separating these axes allows methods from reinforcement learning, imitation learning, and VLA research to be compared without forcing them into a single architectural category.

Our survey scope is defined by function rather than by architecture. We include methods that predict task-relevant future world evolution for manipulation, especially when the prediction is conditioned on robot actions or used to evaluate action consequences. This includes latent dynamics models, action-conditioned video predictors, geometric and spatiotemporal predictors, physics-informed dynamics models, learned simulators, learned environments, and predictive components within VLA systems. We also include datasets and benchmarks that are commonly used to train, evaluate, or stress-test such predictive systems. We exclude pure perception modules, inverse models, policies without explicit prediction, and reward or value functions in isolation, although these components often appear inside systems that use world models. The goal is not to catalog every predictive model in embodied AI, but to identify the design choices that determine whether prediction becomes useful for robotic manipulation.

Our central position is that a world model is best understood not as a particular architecture, modality, or loss function, but as a predictive system whose value depends jointly on what aspects of the future it represents and how those predictions are consumed by robot learning or control. Section~\ref{sec:defining} formalizes this position through an operational definition that distinguishes world models from perception modules, inverse models, policies, reward models, and value functions. Section~\ref{sec:representations} surveys the five representation families and analyzes their trade-offs in fidelity, planning horizon, computational cost, and robustness. Section~\ref{sec:functional} develops the prediction-action taxonomy, while Section~\ref{sec:infra} characterizes world models as learning and decision infrastructure for synthetic experience generation, candidate filtering, search-based evaluation, learned environments, and outcome verification. Section~\ref{sec:lifecycle} reorganizes these roles across pretraining, post-training, and inference adaptation. Section~\ref{sec:datasets} reviews 34 principal datasets, and Section~\ref{sec:benchmarks} synthesizes evaluation protocols spanning predictive fidelity, downstream task performance, and simulator reliability. Together, these perspectives show how world models are evolving from narrow dynamics predictors into predictive infrastructure for robot learning, while clarifying where this expansion improves manipulation and where it mainly shifts the burden to data, verification, or closed-loop evaluation.

\section{Definition and Scope of World Models}
\label{sec:defining}

\begin{figure*}[ht]
    \centering
    \includegraphics[width=\linewidth]{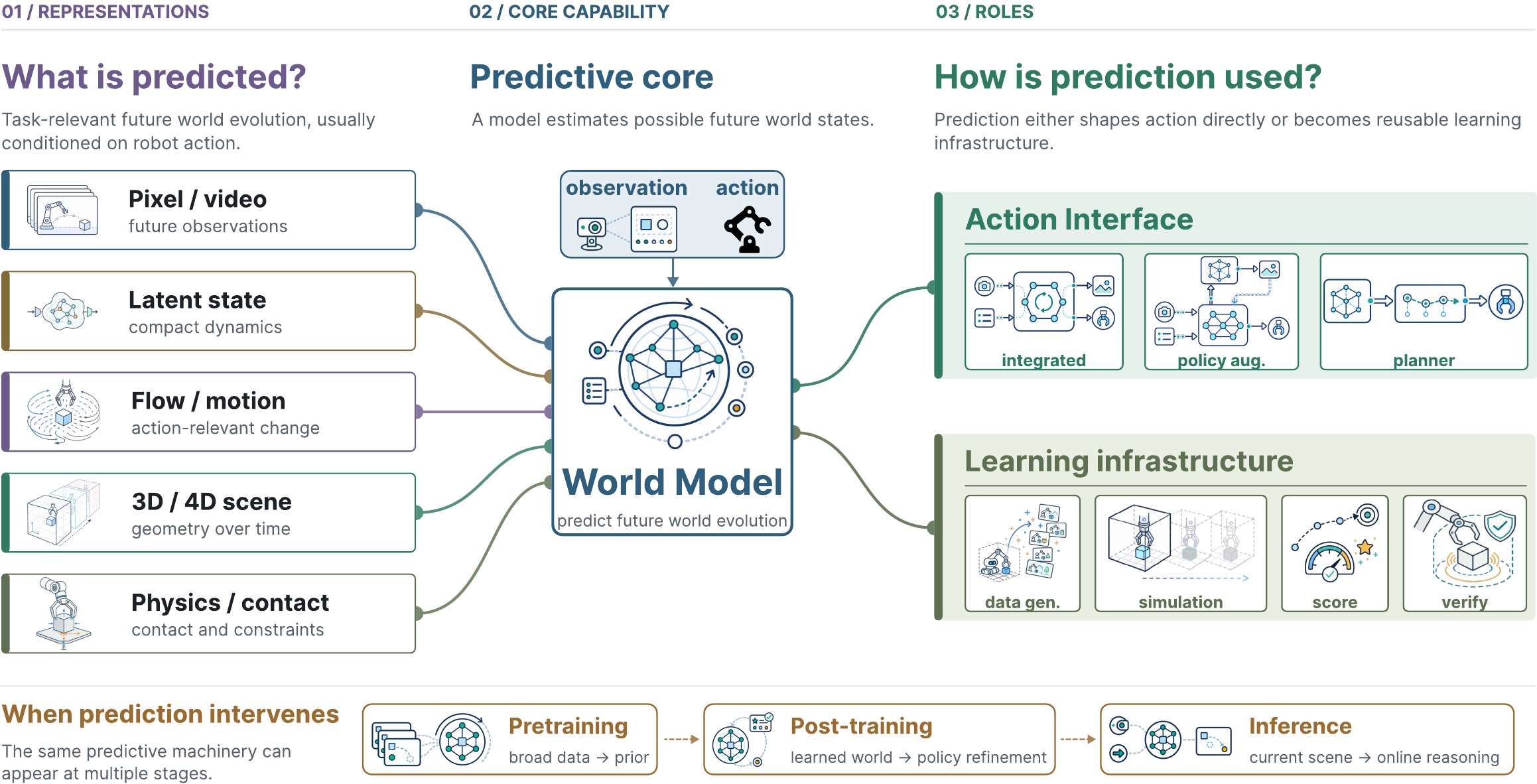}
    \caption{A world model predicts task-relevant future evolution of the world, usually conditioned on observations and robot actions. We organize the literature along three complementary axes: the predicted representation, the predictive core, and the role of prediction in action interfaces, learning infrastructure, and the learning lifecycle.}
    \label{fig:organization}
\end{figure*}

The term \emph{world model} is now used across model-based reinforcement learning, imitation learning, video generation, robotics, and vision-language-action systems, often with different meanings. In one setting, it denotes a compact latent transition model used for planning. In another, it denotes an action-conditioned video generator. In a third, it denotes a learned simulator, a predictive reward source, or a reasoning component inside a generalist policy. A useful definition for robotic manipulation must therefore satisfy two requirements. It must be broad enough to cover the predictive systems that now support robot learning, and narrow enough to exclude perception modules, inverse models, policies, and reward predictors that do not themselves model future world evolution. We begin from the historical roots of the concept in forward models and model-based learning, then define the term operationally for manipulation.

\subsection{From Forward Models to Model-Based Learning}
\label{subsec:wm-history}

Although the label \emph{world model} became widespread only recently, the underlying idea has a long history in cognitive science, motor control, control theory, and reinforcement learning. Earlier work used terms such as \emph{internal model} and \emph{forward model} to describe predictive mechanisms that estimate the sensory or physical consequences of motor commands~\cite{jordan1991internal,jordan1992forward,wolpert1995internal,miall1996forward,wolpert1998paired}. This distinction is central for manipulation. A forward model predicts how the body and environment will change under a candidate action, whereas an inverse model estimates the action needed to reach a desired state. Inverse kinematics and inverse dynamics therefore support command synthesis, while forward models support anticipation, counterfactual comparison, and planning over possible futures.

Model-based reinforcement learning turned this predictive idea into a computational mechanism for decision making. Dyna connected learning, planning, and acting through a learned model of action effects~\cite{sutton1991dyna}. Later probabilistic model-based methods showed that learned dynamics could support policy search under limited interaction budgets~\cite{deisenroth2011lowcost}. Deep model-based reinforcement learning extended the paradigm to high-dimensional observations. PlaNet learned a recurrent latent dynamics model from pixels and used it for online planning, while Dreamer optimized behavior through imagined trajectories in latent space~\cite{planet2018,dreamer2019}. These systems established an important principle that remains relevant for manipulation: a world model need not reconstruct every detail of the world. It is useful when it predicts the aspects of future evolution that a downstream policy, planner, or learning algorithm can exploit.

\subsection{World Models in Contemporary Embodied AI}
\label{subsec:wm-modern}

Recent embodied AI has broadened both the representations and the uses of world models. Ha and Schmidhuber popularized the term in a deep generative setting, where compact latent predictors enabled policy learning inside imagined environments~\cite{ha2018worldmodels}. Since then, the term has expanded beyond latent dynamics for reinforcement learning to include action-conditioned video prediction, learned environments, three- and four-dimensional scene forecasting, physics-informed dynamics, joint world-action models, and predictive components inside VLA systems~\cite{dreamerv3,worldvla,world4rl,worldgym}. This expansion reflects a genuine change in how predictive models are used. They no longer serve only as short-horizon transition models for control. They also provide reusable priors for pretraining, generate synthetic experience, evaluate candidate actions, supply reward or preference signals, support self-correction, and act as learned simulators for policy improvement.

The same expansion also creates ambiguity. Some systems are called world models because they predict visual futures. Others receive the label because they support planning, even when the predicted state is latent and not directly interpretable. Still others are called world models because they provide an environment in which policies can be evaluated or improved. These uses are related but not interchangeable. A video generator that produces plausible futures may fail as a manipulation world model if its predictions are not action-aligned. A latent dynamics model may be effective for control but difficult to inspect for physical plausibility. A learned simulator may be useful for policy ranking but unsafe for policy optimization if it can be exploited. The manipulation setting, therefore, requires a definition based on predictive function rather than on architecture, modality, or training loss.

\subsection{An Operational Definition for Robotic Manipulation}
\label{subsec:wm-definition}

\emph{A world model is a predictive system, learned end-to-end or assembled from learned and analytic components, that estimates how task-relevant aspects of the external world evolve over time and, in robotic manipulation, predicts or evaluates that evolution under robot intervention.}

This definition has three parts. First, the system must be predictive. It must estimate future evolution rather than only encode the present. A static visual encoder, object detector, segmentation model, or language-conditioned perception module may provide inputs to a world model, but it is not itself a world model unless it models temporal change. Second, the system must be world-grounded. It must describe some aspect of the external scene, such as observations, objects, geometry, contacts, physical state, affordances, or latent variables that stand in for these quantities. A reward model, value function, or critic may be useful for action selection, but by itself it summarizes utility rather than modeling how the world changes. Third, for manipulation, the system must be intervention-aware. It must support counterfactual reasoning about what will happen if the robot pushes, grasps, places, inserts, opens, or otherwise acts on the environment. This usually means explicit action conditioning, although some pretrained passive predictors enter our scope when they are adapted, queried, or embedded in systems that evaluate robot action consequences.

The predicted quantity can take many forms. It may be a future image, a sequence of video tokens, a latent state, a motion field, a point cloud, a three- or four-dimensional scene representation, an object trajectory, a contact state, a material deformation, an affordance change, or a physical variable. The representation is secondary to the function. A model qualifies when its predictions support action, learning, simulation, evaluation, verification, or reasoning about future world states. This function-based definition lets us compare methods with substantially different architectures while preserving a clear boundary around predictive world evolution.

This boundary separates world models from neighboring components. Inverse models compute actions from desired states but do not predict the consequences of alternative actions. Policies map observations and goals to actions, but they qualify as world-model components only when they explicitly model future states, observations, or outcomes. Reward models and value functions estimate task utility, but they are not world models in isolation because they abstract away transition structure. Static foundation encoders can make world models more transferable, but they are not world models unless paired with a predictive mechanism. Classical simulators occupy a related but distinct position. They can function as world models when used to predict future environmental evolution, but this survey focuses on learned or hybrid predictive systems because they are the source of the recent conceptual expansion in robot learning.

It is also useful to distinguish three levels at which the term is used. A \emph{world-model component} is a predictive module inside a larger robot system, such as a latent dynamics model, video predictor, geometric predictor, or learned transition model. A \emph{world-model-augmented system} is a policy, planner, or VLA architecture that uses such a component to generate actions, evaluate candidates, or improve learning. A \emph{world-model objective} is a training signal, such as future-frame prediction or latent prediction, that may improve representations even if no explicit rollout is used at deployment. This distinction matters because some recent systems gain most of their benefit from predictive pretraining rather than from test-time imagination. Throughout this survey, we treat these cases separately when possible: predictive components define the scope, augmented systems define the functional interface, and predictive objectives explain how useful representations are learned.

Within this scope, we include latent dynamics models, action-conditioned video predictors, predictive three- and four-dimensional scene models, flow- and motion-based predictors, physics-informed dynamics models, learned simulators, learned environments, and joint world-action models that explicitly model future world evolution. We also include predictive modules in VLA systems that connect language or task goals to future states, observations, action consequences, or outcome verification. We exclude pure perception modules, inverse models, policies without explicit prediction, and reward or value models in isolation. These components remain essential to many robot-learning systems, but they are not world models unless they participate in modeling how the external world evolves.

This operational definition motivates the organization of the rest of the survey. Section~\ref{sec:representations} asks what form of future world evolution is represented. Section~\ref{sec:functional} asks how that prediction is connected to action. Section~\ref{sec:infra} asks how predictive models become reusable learning and decision infrastructure. Section~\ref{sec:lifecycle} asks when the same predictive capability is used during pretraining, post-training, or inference. The definition, therefore, serves not as a fixed architectural category but as a boundary for analyzing how prediction becomes useful in robotic manipulation.

\section{Representations}
\label{sec:representations}

The representation chosen by a world model defines what aspects of the future can be predicted, inspected, optimized, and transferred. For manipulation, this choice is not merely an implementation detail. A pixel-space predictor exposes visually interpretable futures but may spend capacity on texture and lighting. A latent model supports efficient planning but hides the structure that determines physical plausibility. A geometric model improves spatial reasoning but depends on the quality of sensing and reconstruction. A physics-informed model offers stronger extrapolation in principle but requires assumptions about contact, friction, and material parameters that are hard to identify from vision alone. Representation is therefore the first major design axis in world modeling.

\begin{figure*}[ht]
    \centering
    \includegraphics[width=\linewidth]{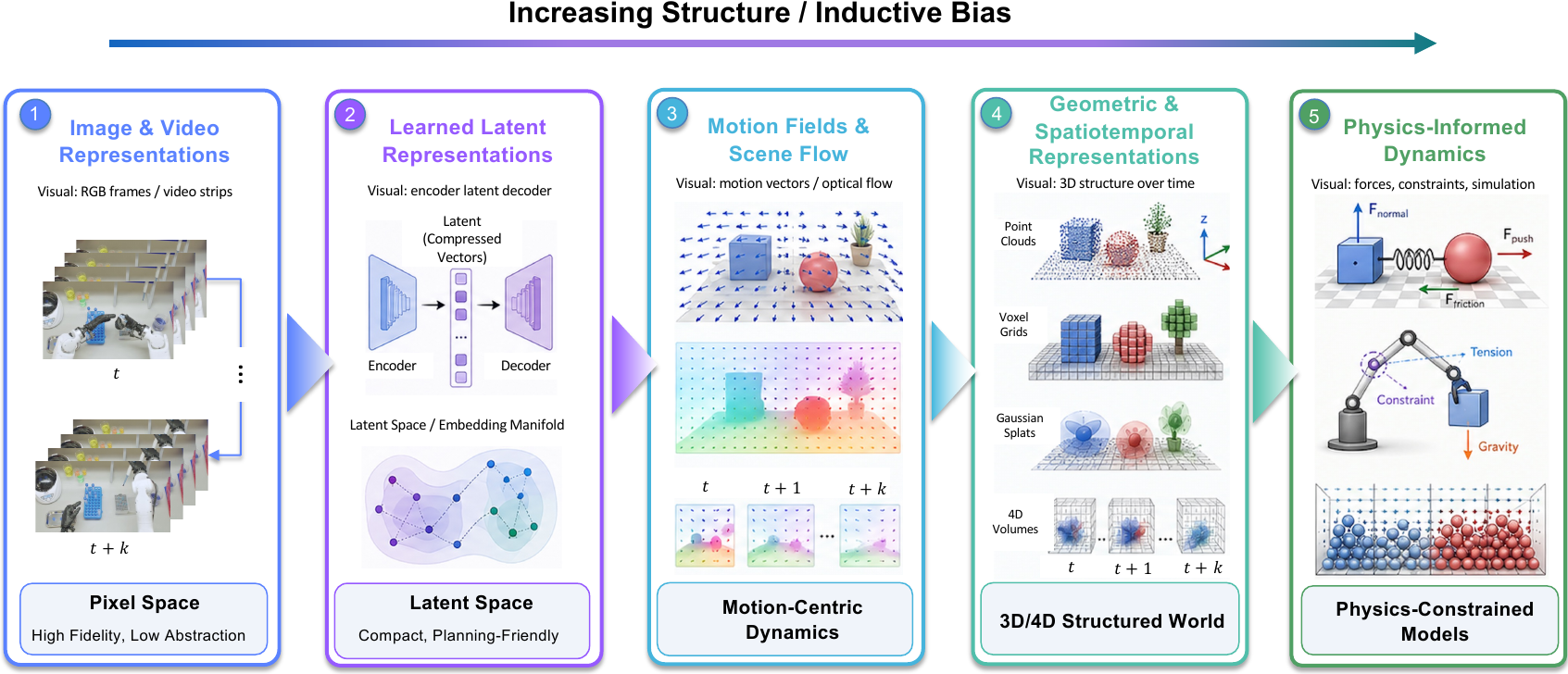}
    \caption{Representation spectrum of world models. The five families are ordered by increasing structured inductive bias, from appearance-reconstructive image and video prediction to physics-informed dynamics. The appropriate representation depends on the downstream role of prediction: visual forecasting, efficient control, motion reasoning, geometric consistency, or physical extrapolation.}
    \label{fig:representation}
\end{figure*}

We organize existing methods into five families: image and video representations, learned latent representations, motion fields and scene flow, geometric and spatiotemporal representations, and physics-informed dynamics. These families differ along two dimensions. The first is what spatial substrate the model predicts, ranging from pixels to latent variables, flow fields, point clouds, volumetric scenes, and physical states. The second is what kind of dynamics the representation makes easy to express, ranging from appearance evolution to task-oriented change and physically constrained transition. Figure~\ref{fig:representation} summarizes this spectrum.

\subsection{Image and Video Representations}

Image and video world models predict directly in the visual domain, using raw frames, image tokens, video tokens, or video-VAE latents. Their main advantage is interface compatibility. Many robot policies already consume images, and a predicted future image or video can serve as a subgoal, a rollout, an imagined environment, or a reasoning trace without requiring an explicit state estimator. UniPi~\cite{unipi} and SuSIE~\cite{susie} showed that video or image generation can be repurposed as a policy substrate, while robotics-oriented systems such as VLP~\cite{vlp}, GR-1~\cite{wu2024unleashing}, and GR-2~\cite{gr2} couple visual prediction with language-conditioned action generation. More recent systems, including DreamGen~\cite{dreamgen}, HMA~\cite{hma}, WorldGym~\cite{worldgym}, Cosmos Policy~\cite{cp}, and Genie Envisioner~\cite{genie_envisioner}, extend this idea toward foundation-scale world modeling, synthetic experience generation, learned environments, and policy evaluation.

The strength of this family is that its predictions are interpretable and naturally aligned with visual imitation (VLA) policies. A generated future can be inspected by a human, scored by a vision-language model, converted into a subgoal, or used as an imagined observation during post-training. This makes image and video representations especially attractive when world models serve as data engines, learned simulators, or explicit visual planners. The weakness is that visual plausibility is not equivalent to action validity. Pixel-space objectives allocate modeling capacity to lighting, texture, background clutter, and camera artifacts, while manipulation success often depends on contacts, object permanence, force closure, and small geometric changes. Rollouts can therefore look coherent while violating the physical constraints that determine whether an action is executable. For this family, the most important evaluations are not only PSNR, SSIM, LPIPS, or FVD, but also action consistency, contact plausibility, and downstream control performance.

\subsection{Learned Latent Representations}

Latent world models compress observations into a learned state space and perform prediction in that compact representation. PlaNet~\cite{planet2018} and Dreamer-style methods~\cite{dreamer2019,dreamerv3} established that latent dynamics can support planning and policy optimization from high-dimensional observations. The key idea is predictive sufficiency. The model does not need to reconstruct every pixel; it needs a state representation that preserves the information required for future reward, action selection, or behavior learning. V-JEPA~2~\cite{vjepa2} pushes this principle further by emphasizing predictive latent learning without pixel reconstruction. In manipulation and VLA systems, WorldVLA~\cite{worldvla}, Fast-WAM~\cite{fast_wam}, LaST-VLA~\cite{last_vla}, Chain of World~\cite{chain_of_world}, and AtomVLA~\cite{atomvla} adopt latent prediction to connect scalable world modeling with action generation, post-training, or reasoning.

The appeal of latent representations lies in their efficiency. Compact latent rollouts are cheaper to generate, search over, and optimize than full visual rollouts. They are therefore well-suited to model-based reinforcement learning, imagined policy learning, candidate evaluation, and inference-time planning under latency constraints. They also allow predictive objectives to shape policy representations even when no explicit rollout is used at deployment. The cost is auditability. A latent state may encode exactly what the training objective needs while discarding the geometric, semantic, or physical variables that a human or verifier would use to judge whether a prediction is plausible. Latent models can also overfit to the task and embodiment distribution on which they were trained. For this family, the decisive question is not whether the latent rollout is visually faithful, but whether it improves downstream success, transfers across tasks and embodiments, and remains calibrated under distribution shift.

\subsection{Motion Fields and Scene Flow}

Motion-centric world models predict displacement rather than full appearance. This family includes optical flow, scene flow, object motion fields, and related spatiotemporal displacement representations. The motivation is that manipulation is often determined by how objects move under intervention rather than by how the entire image changes. FLIP~\cite{flip} represents manipulation futures through flow-centric generative planning, and FlowVLA~\cite{flowvla} extends this idea to language-conditioned motion reasoning for VLA systems. By focusing prediction on motion, these models reduce the burden of photorealistic synthesis and expose a signal that is closer to action consequences.

The strength of flow representations is their alignment with short-horizon physical change. They can highlight which parts of the scene move, how object surfaces deform or translate, and whether a candidate action produces progress toward a goal. This makes them useful for pushing, rearrangement, tool use, and other tasks where the key uncertainty lies in object displacement. Their weakness is that motion alone is incomplete. Flow fields may discard object identity, semantic attributes, texture cues, and static scene constraints that matter for language-conditioned manipulation. They can also be ambiguous in scenes where contact has not yet occurred, since the most important event is a future discontinuity rather than a current motion pattern. Flow-based world models are therefore strongest when paired with semantic or geometric representations that recover what motion fields intentionally omit.

\subsection{Geometric and Spatiotemporal Representations}

Geometric world models represent the scene through depth, point clouds, object-centric structure, 3D features, neural fields, Gaussian splats, or four-dimensional spatiotemporal volumes. Their motivation is that many manipulation failures are spatial in nature. A two-dimensional image can hide occluded objects, confuse viewpoint changes with object motion, or make it difficult to infer contact geometry. 3D-VLA~\cite{3d_vla} and OG-VLA~\cite{og_vla} showed that grounding VLA systems in geometric structure can improve generalization under viewpoint variation. 3D-CAVLA~\cite{3d_cavla} and PointWorld~\cite{pointworld} further emphasize point-cloud and 3D context for scene understanding and future prediction. TesserAct~\cite{tesseract} and WristWorld~\cite{wristworld} extend the representation to four-dimensional scene evolution, while GWM~\cite{lu2025gwm} uses Gaussian-based scene modeling to connect differentiable rendering with manipulation world models.

The main advantage of geometric representations is spatial consistency. They make occlusion, viewpoint change, object displacement, and robot-scene geometry more explicit than image-only prediction. This is important for grasping, placing, insertion, assembly, and long-horizon rearrangement, where success depends on spatial relations that may be difficult to infer from a single RGB view. Geometry also makes certain failures easier to detect, such as object penetration, implausible motion, or inconsistent multi-view predictions. The cost is infrastructure. Accurate depth, multi-view capture, point-cloud reconstruction, or volumetric scene modeling is less available than RGB video at the internet scale. Four-dimensional representations can be computationally expensive, and their benefits are most evident in tabletop settings where spatial scope is limited. Geometric world models are therefore most compelling when spatial precision and viewpoint robustness matter more than raw data scale.

\subsection{Physics-Informed Dynamics}

Physics-informed world models encode or impose constraints from rigid-body dynamics, contact mechanics, friction, deformation, or material behavior. They sit at the high-inductive-bias end of the representation spectrum. PIN-WM~\cite{li2025pinwm} illustrates this direction by incorporating physics-informed priors for non-prehensile manipulation, where action outcomes depend strongly on contact and object dynamics. Related systems combine learned prediction with differentiable simulation, physical priors, or structured dynamic variables so that generated futures obey constraints that purely statistical predictors often violate.

The strength of physics-informed representations is extrapolation under the right assumptions. When object geometry, material properties, and contact parameters are known or identifiable, physical structure can improve robustness outside the training distribution and reduce hallucinated transitions. This is especially valuable for pushing, sliding, insertion, deformable-object manipulation, and tool-mediated interaction, where small physical errors can determine success or failure. The weakness is that manipulation often violates the assumptions required by clean physical modeling. Contact is discontinuous, friction is hard to estimate, deformable materials are difficult to parameterize, and real scenes contain unmodeled compliance, perception error, and actuator uncertainty. Differentiable physics can therefore be brittle exactly where manipulation is most interesting. Physics-informed world models are best viewed not as replacements for learned representations, but as constraints or priors that can improve reliability when paired with data-driven perception and verification.

\subsection{Discussion: Choosing a Representation}
\label{sec:representation-discussion}

The five representation families differ mainly in where they place the burden of generalization. Image and video models place it on the data scale and generative capacity. Latent models embed it into the training objective and the downstream policy. Motion-field models assume that displacement captures the action-relevant future. Geometric models place it on sensing and reconstruction. Physics-informed models place it on system identification and the validity of physical assumptions. No representation dominates across all manipulation settings.

A useful design rule is to choose the representation based on the world model's role. If the model supplies visual subgoals, synthetic experience, or a learned environment for a visual policy, image and video representations are natural, but they require explicit checks for action consistency. If the model is used for fast imagined rollouts, policy learning, or model predictive control, latent representations are usually more efficient, but they require downstream and out-of-distribution evaluation. If the task depends on short-horizon object displacement, flow representations can directly expose the relevant change. If viewpoint, occlusion, or spatial precision dominates, geometric and spatiotemporal representations are preferable. If contact, deformation, or non-prehensile dynamics dominate, physics-informed structure becomes valuable, but only when paired with reliable estimation and verification.

The current trend is therefore hybrid rather than exclusive. Video models are being conditioned on flow, latent predictors are being built over geometric primitives, Gaussian scene models are being combined with physical constraints, and learned simulators are being augmented with verifiers. This convergence suggests that representation should be treated as a compositional design choice. A manipulation world model rarely needs to predict everything. It needs to predict the future variables that determine whether a robot action will remain executable, physically credible, and useful for the policy or planner that consumes it.

\section{A Functional Taxonomy by Prediction--Action Interface}
\label{sec:functional}

Representations describe what a world model predicts. A separate question is how that prediction changes robot actions. The same predicted future can play very different roles: it may be hidden inside a policy, exposed as a subgoal, decoded into actions through inverse dynamics, searched over by a planner, or used only as an auxiliary training signal. We therefore organize this section by the direct interface between prediction and action generation. This view cuts across reinforcement learning, imitation learning, and VLA systems, because pixel, latent, flow, and geometric predictors can all support either integrated policies or explicit planners~\cite{worldvla,cot_vla,vjepa2,flip,tan_vla_survey}.

The key distinction is whether prediction is internal or exposed. In \emph{integrated prediction--action models}, future modeling is fused with the action-producing model, and the predicted future may never appear as a separate object at deployment. In \emph{explicit predictive planners}, the world model exposes an intermediate target, such as a subgoal image, a rollout, a waypoint sequence, or a latent plan, that another controller, inverse model, or optimizer must realize. This section focuses on direct-action interfaces, in which prediction participates in producing the next action or plan. Section~\ref{sec:infra} considers a broader systems role, where world models act as reusable infrastructure for data generation, scoring, simulation, search, or verification across many candidate futures.

\begin{figure*}[t]
    \centering
    \includegraphics[width=\textwidth]{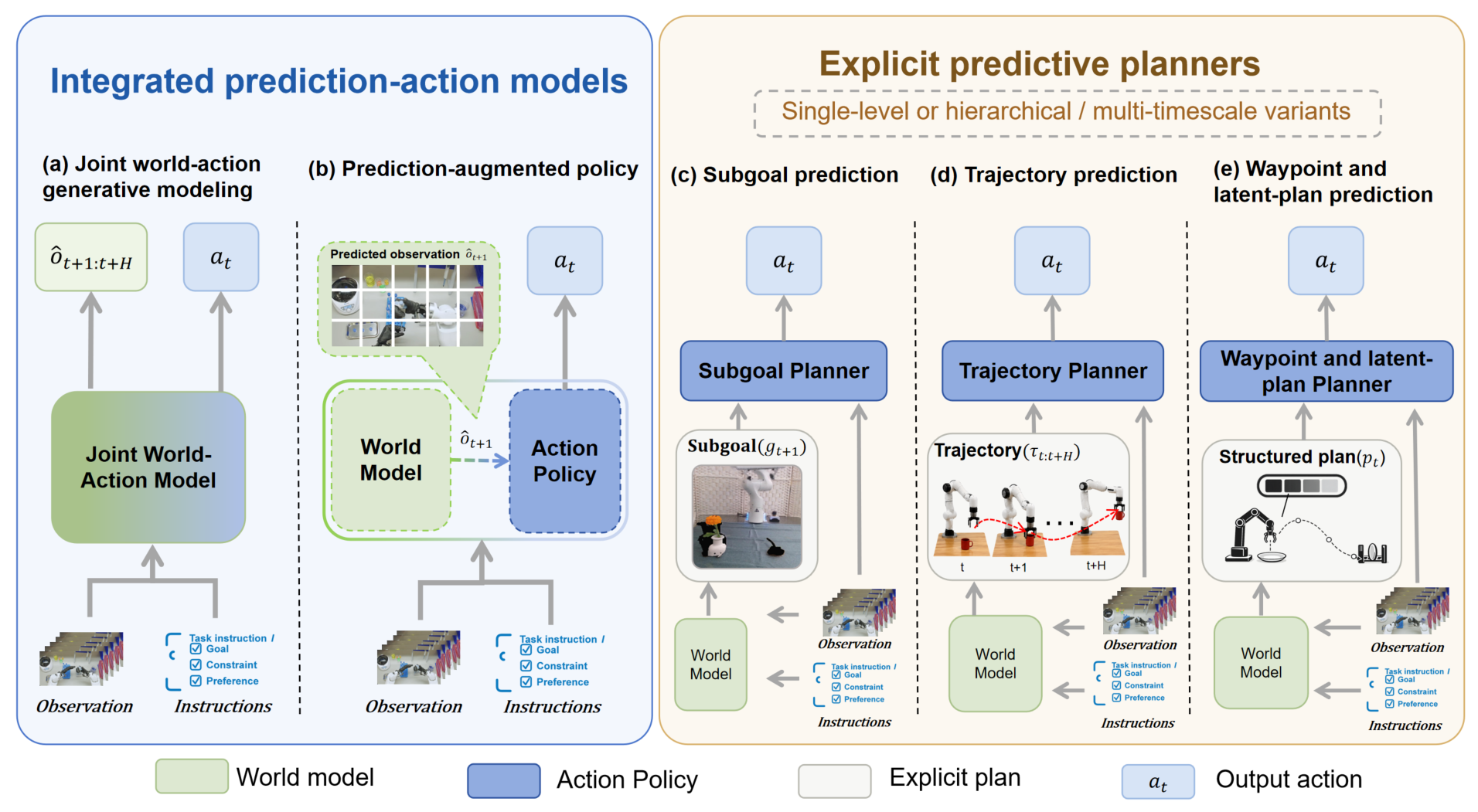}
    \caption{
    Functional taxonomy of direct prediction--action interfaces. (a--b) Integrated prediction--action models embed prediction inside the action-producing model. (c--e) Explicit predictive planners expose prediction as an intermediate target---a subgoal $g_{t+1}$, a trajectory $\tau_{t:t+H}$, or a compact structured plan $p_t$ (waypoints or latent plan)---that a downstream controller must realize. Hierarchical and multi-timescale variants apply orthogonally to subgoal, trajectory, and structured-plan interfaces. Here $o_t$ denotes the current observation, $a_t$ the action at time $t$, and $\hat{o}_{t+1}$ and $\hat{o}_{t+1:t+H}$ short- and longer-horizon future predictions.
    }
    \label{fig:Functional_taxonomy}
\end{figure*}

\subsection{Integrated Prediction--Action Models}

Integrated prediction--action models embed predictive structure inside the controller itself. The model may predict future frames, latent states, action tokens, motion features, or semantic dynamics during training, but the deployed system typically outputs actions directly. This interface is attractive because it avoids a brittle handoff between a world model and a separate executor. The policy can learn representations that are useful for both prediction and control, and prediction can regularize action generation toward physically grounded behavior. The cost is interpretability. When prediction is internal, it becomes harder to inspect whether the policy is acting because it has learned useful dynamics or because the predictive loss merely improved representation learning.

\noindent\textbf{Joint world--action generative modeling.}
The strongest form of integration treats future observations and actions as one generative sequence. GR-1, GR-2, and HMA learn joint visual and action dynamics through autoregressive token modeling~\cite{wu2024unleashing,gr2,hma}. WorldVLA and RynnVLA-002 extend this idea to multimodal sequence models in which image, language, and action streams share a predictive backbone~\cite{worldvla,rynnvla002}. PAR introduces more physically grounded tokenization, while diffusion-based variants such as DUST and zero-shot world-action modeling aim to improve continuous control fidelity beyond pure autoregression~\cite{par,dust,wam_zero_shot}. The central advantage is consistency between what the model imagines and what it can execute. Because future prediction and action generation are jointly optimized, the model is encouraged to represent action consequences rather than merely passive visual continuation. The central limitation is the scaling cost. Joint world--action models require large multimodal datasets, high-capacity sequence models, and careful balancing between visual prediction and control accuracy. Recent analyses further suggest that the benefit of this family may come as much from predictive representation learning as from explicit test-time imagination~\cite{fast_wam,wam_generalization}.

\noindent\textbf{Prediction-augmented policy learning.}
A looser form of integration keeps the architecture policy-centric while using prediction as an auxiliary signal, bottleneck, or reasoning trace. In imitation learning, Seer aligns foresight tokens with action tokens, while FLARE replaces explicit frame generation with future-latent alignment~\cite{seer,flare}. In VLA systems, CoT-VLA inserts predicted future frames as visual reasoning steps, DreamVLA predicts dynamic and semantic world knowledge, and VLA-JEPA, FlowVLA, and 3D-VLA move prediction into latent, motion, or geometric spaces that are closer to control-relevant structure~\cite{cot_vla,dreamvla,vla_jepa,flowvla,3d_vla}. DIAL and UP-VLA use predictive or intent-like bottlenecks between high-level instruction understanding and low-level action generation~\cite{dial,up_vla}. Across these systems, prediction shapes the policy without necessarily becoming a separately optimized planner.

The same principle appears in imagination-based policy learning. Dreamer-style methods optimize policies through latent rollouts~\cite{dreamerv3}, and DayDreamer demonstrated that this paradigm can transfer to physical robots~\cite{wu2023daydreamer}. Later manipulation systems strengthen the predictive substrate with multi-view consistency, view-invariant prediction, or contact-sensitive latent diffusion, as in Multi-View Masked World Models, ReViWo, and LaDi-WM~\cite{seo2023mvwm,pang2025reviwo,huang2025ladiwm}. LUMOS trains language-conditioned skills through latent practice, IQ-MPC combines a world model, critic, and policy within inverse-soft-Q imitation, and FOCUS and SeeX use internal prediction to improve data collection or exploration~\cite{lumos,iqmpc,ferraro2025focus,huang2024seex}. The benefit is that policies receive supervision about futures that are absent from demonstrations or sparse rewards. The limitation is weaker causal accountability. Unless the predictive pathway is carefully exposed or ablated, improved performance may reflect better representation learning rather than reliable online reasoning.

\subsection{Explicit Predictive Planners}

Explicit predictive planners expose prediction as an intermediate object that another module must execute. The predicted object can be a goal image, a future video, a latent trajectory, a waypoint sequence, or a hierarchy of subgoals. This interface is easier to inspect than integrated prediction because the plan can often be visualized, scored, or replaced. It also allows modular reuse: a strong world model can be paired with different controllers, and a strong controller can execute different predicted targets. The weakness is the handoff problem. A predicted future may be desirable but unreachable, and a controller may fail even when the prediction is correct. Explicit planning, therefore, shifts the burden from representation learning to executability and closed-loop correction.

\noindent\textbf{Subgoal prediction.}
The simplest explicit interface predicts a single future target. SuSIE uses an image-editing diffusion model to generate subgoal images that a goal-conditioned controller executes~\cite{susie}. GR-MG adapts the same image-level interface to partially annotated corpora, and Imagine2Act enriches subgoals with 3D point-cloud information for finer spatial precision~\cite{grmg,imagine2act}. MinD uses a related dual-system design, pairing a low-frequency visual planner with a high-frequency controller~\cite{mind}. Subgoal prediction is attractive because it exposes an interpretable target and keeps the inner control loop fast. Its limitation is temporal underspecification. A single goal image says little about the contact sequence, approach path, or recovery behavior needed to reach it, which makes the interface brittle for long-horizon or contact-rich tasks.

\noindent\textbf{Trajectory prediction.}
Trajectory planners predict a short sequence of future states rather than a single target. UniPi generates text-conditioned future videos and decodes actions through inverse dynamics~\cite{unipi}. CLOVER closes the loop by replanning when execution drifts from the predicted trajectory, and EVA improves executability by rewarding rollouts that remain compatible with action decoding~\cite{clover,eva_il}. Latent variants follow the same interface with lower rollout cost. V-JEPA~2 predicts future latent states from large-scale video pretraining, and VPP uses predictive visual trajectories to guide inverse-dynamics control~\cite{vjepa2,vpp}. In reinforcement learning, TD-MPC2 and the MoDem line use short latent rollouts or demonstration-augmented model predictive control to choose actions~\cite{hansen2024tdmpc2,hansen2022modem,lancaster2024modemv2,lopezescoriza2025demo3}. Trajectory prediction provides richer temporal guidance than subgoal prediction, but it is more exposed to compounding error. The longer the predicted sequence, the more the planner must handle drift between imagined and realized states.

\noindent\textbf{Waypoint and latent-plan prediction.}
Some planners compress the future into structured intermediate variables rather than full observations. PIVOT-R predicts task-relevant waypoints for downstream execution, while PIN-WM uses a physics-informed differentiable world model to optimize contact-rich manipulation plans~\cite{pivot_r,li2025pinwm}. More broadly, many latent planning methods fit this category when the decision variable is a compact plan optimized at inference time rather than an explicit image or video. The advantage is search efficiency. A waypoint or latent plan can preserve the action-relevant structure of a future while avoiding the cost of high-dimensional visual rollout. The disadvantage is reduced transparency. Latent plans are efficient partly because they hide details, making them harder to verify, debug, and communicate across embodiments.

\noindent\textbf{Hierarchical predictive planning.}
Hierarchy is not a separate plan type, but a way of stacking the previous interfaces across temporal scales. Hierarchical Planning with Latent World Models combines multi-timescale latent planning with real-robot execution~\cite{zhang2026hwm}. VLP combines language subgoals with text-to-video generation, Reflective Planning revises long-horizon decisions through imagined futures, NovaPlan couples video-language planning with geometric execution, RoboHorizon uses LLM-guided task decomposition, and TriVLA separates grounding, episodic world modeling, and low-level control into coordinated subsystems~\cite{vlp,reflective_planning,novaplan,robohorizon,trivla}. Hierarchy helps because long-horizon manipulation is rarely solved by one uniform prediction horizon. High-level predictions determine what should happen next, while low-level predictions assess local feasibility. The risk is error propagation. A wrong high-level subgoal can make lower-level prediction and control appear competent while still driving the system toward failure.

\subsection{Design Implications}

The direct prediction--action interface suggests a practical classification rule. If prediction is fused into the action-producing model and is not exposed as a separate target at deployment, the method is integrated. If prediction is exposed as a subgoal, trajectory, waypoint sequence, latent plan, or hierarchy that another module must realize, the method is an explicit predictive planner. If the world model is instead used to generate data, score candidates, provide a learned environment, rank policies, or verify outcomes across many possible futures, its dominant role is as infrastructure, as discussed in Section~\ref{sec:infra}.

The choice between integrated and explicit interfaces is a trade-off between coupling and inspectability. Integrated models avoid modular mismatch and can learn prediction and action jointly, but their internal futures are difficult to audit. Explicit planners expose the predicted future and support modular control, but they must solve executability, replanning, and error propagation. This trade-off explains why current systems increasingly blur the boundary: integrated VLA policies add traces of predictive reasoning, explicit planners add learned inverse dynamics and closed-loop correction, and infrastructure models provide external verification for both. For manipulation, the most reliable systems are likely to be those that combine tight action coupling with some exposed mechanism for checking whether the imagined future remains physically reachable.

\section{World Models as Learning and Decision Infrastructure}
\label{sec:infra}

\begin{figure*}
    \centering
    \includegraphics[width=1.0\linewidth]{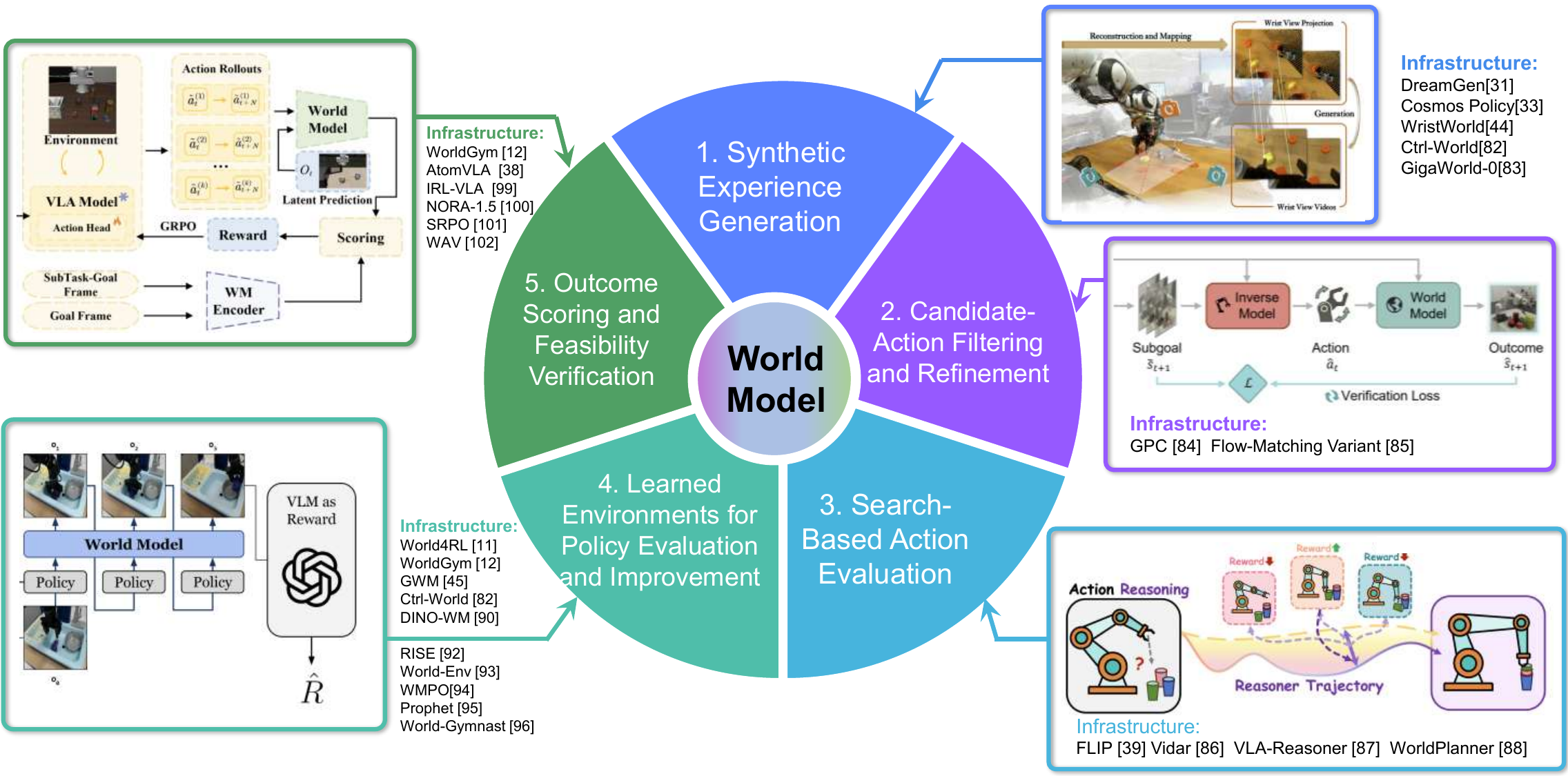}
    \caption{Five functional roles of infrastructure world models for robotic manipulation: synthetic experience generation, candidate-action filtering, search-based evaluation, learned environments, and outcome scoring or verification. The roles are complementary and often combined within a single pipeline; representative methods are listed beside each wedge.}
    \label{fig:infra-roles}
\end{figure*}

World models increasingly serve as infrastructure for robot learning and decision-making rather than merely as components of a single action-generation pathway. The distinction from the prediction--action taxonomy in Section~\ref{sec:functional} is the unit of use. There, prediction is internal to a policy or exposed as a target that guides one decision sequence. Here, the same predictive capability is reused across many candidates, rollouts, training examples, policy updates, reward estimates, or verification decisions. In this infrastructure role, the world model may still affect action selection, but its value lies in being repeatedly queried as a data engine, evaluator, simulator, or judge.

This systems-level role is especially important for robotic manipulation because real interaction is expensive, evaluation can be risky, and many physical tasks are slow, irreversible, or difficult to reset. A world model used as infrastructure can expand the data available for learning, filter actions proposed by an existing policy, support counterfactual search, provide an imagined environment for policy improvement, or determine whether predicted outcomes are trustworthy. We organize these uses by the bottleneck each role addresses, as summarized in Figure~\ref{fig:infra-roles}. Synthetic experience generation aims to address limited coverage in demonstrations and robot data. Candidate-action filtering targets imperfect behavior from an existing policy. Search-based evaluation targets online counterfactual decision making. Learned environments target the cost of policy evaluation and improvement. Outcome scoring and verification target the reliability of imagined experience. These roles are functional rather than mutually exclusive, and a single system may combine several of them.

\subsection{Synthetic Experience Generation}

Synthetic experience generation uses a world model to enlarge the distribution from which policies learn. Manipulation policies often fail not because the learning algorithm is weak, but because the available demonstrations cover too few objects, viewpoints, task variations, contact events, or recovery behaviors. A generative world model can address this bottleneck before reinforcement learning or online adaptation begins by producing future observations, alternative views, full trajectories, or action-conditioned outcomes that expand the effective training set.

DreamGen made this data-engine role explicit by using video generation to synthesize photorealistic manipulation trajectories and then recovering pseudo-actions to construct neural trajectories for downstream policy learning~\cite{dreamgen}. Its contribution was not simply to generate plausible robot videos, but to turn generated futures into supervision that a policy could consume. Subsequent work extended this template in several directions. Ctrl-World emphasizes controllable synthetic generation, reflecting the need for generated trajectories to be conditionable rather than only diverse~\cite{guo2025ctrlworld}. WristWorld targets a specific perceptual gap by generating wrist-view and four-dimensional manipulation observations that are difficult to collect systematically in real environments~\cite{wristworld}. GigaWorld-0 frames world models as scalable data engines for embodied learning, while Cosmos Policy pushes toward foundation-scale generation of action-conditioned futures for visuomotor control and planning~\cite{gigaworld-0,cp}.

The main failure mode is action inconsistency. Generated trajectories can broaden visual diversity while injecting unexecutable or physically inconsistent supervision. A video may depict a successful grasp, insertion, or rearrangement without encoding the forces, contacts, or intermediate actions required to realize it. Synthetic experience, therefore, requires controllability, action annotations, pseudo-action recovery, or downstream filtering to avoid training policies on attractive but unreachable futures. This role is strongest when generation expands coverage of long-tail variations and recovery behaviors while remaining aligned with robot embodiment and action feasibility.

\subsection{Candidate-Action Filtering and Refinement}

Candidate-action filtering uses a world model as a predictive critic around an existing policy. The bottleneck is not the absence of a policy, but the unreliability of its proposals under distribution shift, clutter, contact uncertainty, or long-horizon execution. The policy proposes actions, action chunks, or trajectories, and the world model filters, reranks, or refines them according to predicted feasibility, progress, safety, or task success. This interface is attractive because it can improve a strong pretrained policy without replacing the policy itself.

Generative Predictive Control illustrates this pattern by pairing a frozen diffusion behavior-cloning policy with an independently trained action-conditioned video world model. The policy proposes short-horizon actions, and the world model reranks them based on predicted feasibility on real manipulation systems~\cite{gpc_diffusion}. A flow-matching variant extends the same idea to higher-frequency control with test-time warm starts~\cite{gpc_flow}. In both cases, the world model acts as an external check on proposed behavior rather than as the primary action generator. This makes filtering practical for deployment because it preserves the diversity and fluency of the base policy while adding a predictive layer that can reject likely failures.

The limitation is proposal dependence. Candidate filtering can only improve the actions it receives. If the base policy never proposes a successful action, scoring alone cannot recover one. Performance, therefore, depends on both proposal diversity and score calibration. A predicted future may look plausible while being dynamically impossible, and a high score may reflect visual bias rather than real task progress. Candidate filtering is therefore best viewed as a deployment-time refinement mechanism whose reliability depends on whether the world-model score correlates with real feasibility and task success.

\subsection{Search-Based Action Evaluation}

Search-based action evaluation strengthens candidate filtering by placing the world model inside a repeated decision procedure. The bottleneck is online counterfactual comparison. Instead of scoring a small batch of proposals once, the system queries the model many times as it expands a search tree, optimizes a latent trajectory, or performs model predictive control. The world model serves as an evaluation oracle that estimates what may happen under alternative decisions before the robot commits to one.

FLIP and Vidar predict future motion or video under candidate actions and use these predicted outcomes to guide control~\cite{flip,vidar}. VLA-Reasoner repeatedly queries a world model inside Monte Carlo tree search to compare candidate trajectories online~\cite{vla_reasoner}. WorldPlanner applies a related idea to imitation-style manipulation by combining a visual world model with Monte Carlo tree search over a diffusion action prior, while $\text{M}^{3}\text{PC}$ shows how a pretrained masked trajectory model can support inference-time model predictive control through repeated latent evaluation~\cite{worldplanner,wen2025m3pc}. Across these methods, the key advantage is counterfactual selection: the robot can compare multiple imagined futures before committing to the next action.

This role differs from explicit predictive planning because the model is used primarily as an evaluator within a decision algorithm, rather than as the producer of an exposed plan. Its predictions may never be handed to a separate executor as a subgoal or trajectory. The output provides evidence for choosing among candidates. The main failure modes are latency and over-optimization of model error. Search can amplify the usefulness of a calibrated world model, but it can also amplify bias because the decision procedure repeatedly queries and optimizes against the same imperfect predictor. High-dimensional video or 3D rollouts make this especially costly, which is why latent dynamics models such as M$^3$PC and DINO-WM are attractive substrates for this role~\cite{wen2025m3pc,zhou2025dinowm}. Search-based evaluation is therefore most useful when the robustness gained from counterfactual comparison justifies the added inference cost and when model errors are controlled by replanning, uncertainty estimates, or verification.

\subsection{Learned Environments for Policy Evaluation and Improvement}

A learned environment is the most decoupled infrastructure role. The bottleneck is the cost and risk of evaluating or improving policies through real-world interaction. In this role, the world model becomes a virtual environment in which policies can be rolled out, compared, refined, or stress-tested. This differs from synthetic data generation because the imagined environment can support closed-loop interaction and policy optimization, not only offline supervision.

Some learned environments are compact and state-centric. Robotic World Model and DINO-WM use learned feature dynamics as optimization surfaces for control and policy refinement~\cite{li2025roboticworldmodel,zhou2025dinowm}. Others synthesize observations directly. World4RL refines policies inside a diffusion world model, RISE moves toward self-improving policy and world-model loops, GWM uses three-dimensional Gaussian representations for geometry-aware scene evolution, and Ctrl-World uses controllable generation to synthesize success trajectories that improve downstream behavior~\cite{world4rl,li2025rise,lu2025gwm,guo2025ctrlworld}. State-centric environments are cheaper to roll out and easier to optimize. Observation-centric environments align more naturally with visual policies, but they are slower and more vulnerable to visually plausible yet physically inconsistent artifacts.

A recent line treats learned worlds as reinforcement-learning substrates for VLA post-training. WorldGym frames the world model as an environment for policy evaluation by combining action-conditioned rollouts with VLM-based scoring to assess candidate policies in imagination before real deployment~\cite{worldgym}. World-Env moves from evaluation to reinforcement-style post-training under extreme data scarcity by adding a VLM-guided reflector that supplies dense rewards and predicts task termination~\cite{world_env}. WMPO and Prophet further treat imagined interaction as an optimization substrate for action policies, while World-Gymnast shows that reinforcement learning inside an action-conditioned video world model can improve manipulation policies beyond supervised fine-tuning and beyond reinforcement learning in conventional software simulators~\cite{wmpo,prophet,world-gymnast}. World-VLA-Loop closes the loop by feeding policy failures back into world-model learning, and WoVR focuses on hallucination, long-horizon drift, and policy exploitation of simulator inaccuracies~\cite{world-vla-loop,wovr}.

The central risk is simulator exploitation. A policy trained in a learned environment may discover behaviors that exploit model artifacts rather than transferable manipulation skills. This risk is more severe than ordinary prediction error because policy optimization actively searches for weaknesses in the simulator. A learned environment should therefore be evaluated not only by visual quality or rollout likelihood, but by whether it preserves policy rankings, supports real-world improvement, and avoids exploitable hallucinations under closed-loop training.

\subsection{Outcome Scoring and Feasibility Verification}

Outcome scoring converts imagined futures into quantities that can guide learning or decision making, such as rewards, progress estimates, preferences, feasibility judgments, or safety scores. The bottleneck is feedback. Many manipulation tasks provide sparse rewards or delayed success signals, and generated rollouts are useful only if the system can decide which imagined outcomes represent progress. This role often appears alongside learned environments, but it can also operate independently when the world model provides feedback without serving as a full simulator.

WorldGym instantiates this role by using VLM-based reward scoring to evaluate rollout outcomes in the world model~\cite{worldgym}. IRL-VLA makes the idea more explicit through a reward world model, highlighting that world modeling can be useful not only for generating future observations but also for supplying optimization signals for VLA learning~\cite{irl-vla}. NORA-1.5 uses world-model-informed preference rewards for DPO-style adaptation, SRPO uses world-model latent representations as a progress metric for dense self-referential rewards, and AtomVLA uses a predictive latent world model to score candidate action chunks against decomposed subtask goals during offline post-training~\cite{nora15,srpo,atomvla}. In these systems, the world model changes role from predictor to judge. Its value lies in turning imagined trajectories into feedback that policies can optimize against.

Verification is stricter than scoring. Scoring asks how good an imagined outcome is, whereas verification asks whether the imagined transition should be trusted at all. World Action Verifier extends the judge interface by assessing whether predicted transitions are plausible and reachable~\cite{wav}. By decomposing verification into state plausibility and action reachability, it addresses a central weakness of simulator-centric world modeling: realistic-looking imagined futures are not necessarily feasible. Verification can detect model errors, filter imagined experiences, and support self-improvement in underexplored regimes. Its own failure mode is shared blindness. A verifier trained on the same data or representation as the generator may miss the same contact, geometry, or action-feasibility errors. Reliable verification, therefore, benefits from independent signals, explicit physical constraints, or real-world calibration.

\subsection{Design Implications}

The infrastructure view changes how world models should be evaluated. A model used as a data engine should be judged by whether the experience it generates improves real policy performance, not just by whether its videos look realistic. A model used for filtering or search should be judged by calibration and action alignment, not only by rollout likelihood. A learned environment should be judged by whether it preserves policy rankings and supports improvement without exploitable artifacts. A verifier should be judged by whether it detects failures that the generator itself misses. Across these roles, the central question is not whether the world model can imagine a plausible future, but whether the imagined future remains useful in closed-loop policy learning and decision-making.

These infrastructure roles show why world models are becoming reusable substrates for robot learning. They expand experience before training, improve decisions around existing policies, provide environments for post-training, and judge whether imagined futures can be trusted. The same predictive substrate can recur across pretraining, post-training, and inference, with different failure modes at each stage. This systems perspective complements the direct action-interface taxonomy and motivates the lifecycle view developed next.

\begin{figure*}[htbp]
  \centering
  \includegraphics[width=\textwidth]{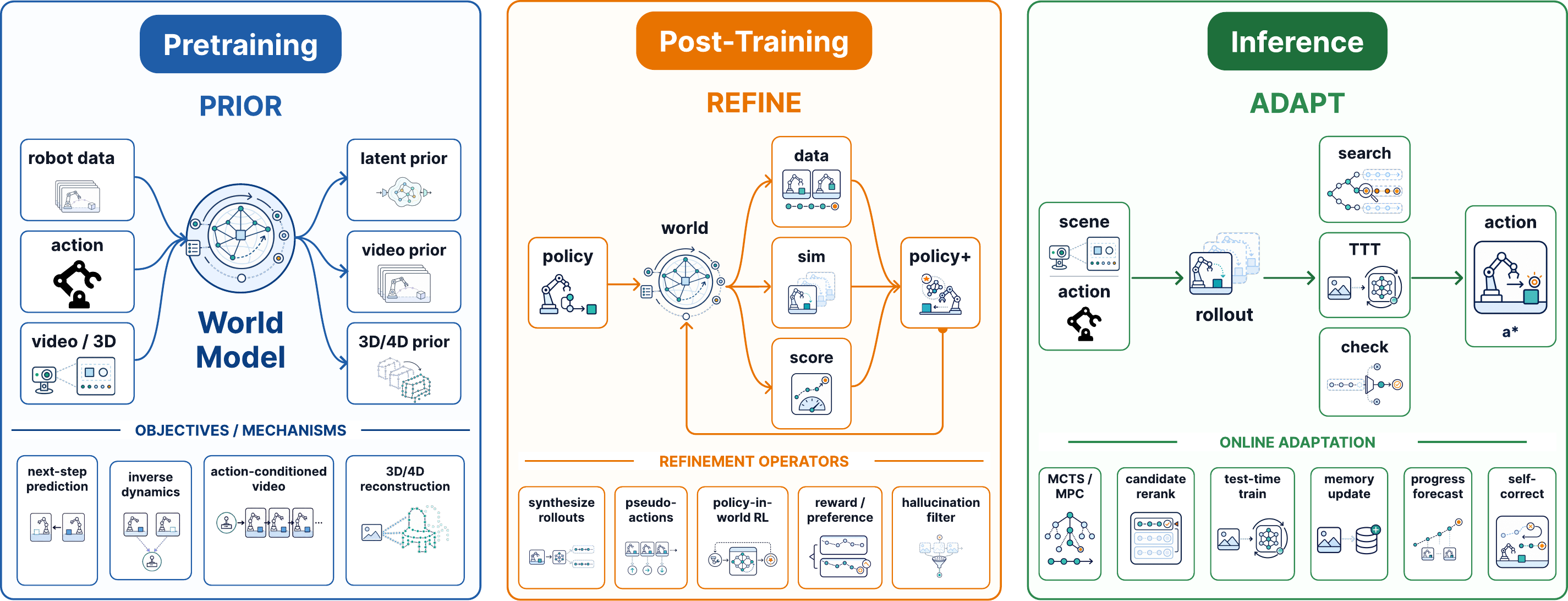}
  \caption{World models across the robot-learning lifecycle. During pretraining, predictive objectives learn reusable latent, video, or three-dimensional priors from robot data, actions, and visual observations. During post-training, the learned world supports policy refinement through synthetic data, simulation, reward or preference signals, and hallucination filtering. At inference time, prediction becomes an online adaptation mechanism for search, candidate reranking, test-time training, memory update, progress forecasting, and self-correction.}
  \label{fig:lifecycle}
\end{figure*}

\section{World Models Across the Learning Lifecycle}
\label{sec:lifecycle}

\begin{table*}[htbp]
  \centering
  \caption{Representative world-model methods categorized across the survey taxonomy. Each row reports the role most emphasized by the method, together with its learning context, representation family, predicted signal, predictive function, and lifecycle stage. The categories are not mutually exclusive: many systems use one predictive model across multiple stages or combine several roles within a single pipeline. Stage abbreviations: \emph{Pre} = pretraining; \emph{Train} = joint policy and world-model training; \emph{Post} = post-training; \emph{Infer} = inference-time use; \emph{Multi} = spans multiple stages.}
  \label{tab:integrated_wm_roles}
  \scriptsize
  \renewcommand{\arraystretch}{1.02}
  \setlength{\tabcolsep}{2.1pt}
  \begin{tabular*}{\textwidth}{@{\extracolsep{\fill}}p{1.2cm} p{2.1cm} p{0.6cm} p{1.05cm} p{1.95cm} p{2.7cm} p{0.65cm}@{}}
    \toprule
    \shortstack[c]{Functional\\role} & Method & Context & Representation & Predicted signal & Predictive function & Stage \\
    \midrule
    \multirow[c]{4}{1.35cm}[-0.4ex]{\centering\shortstack[c]{Predictive\\prior}}
      & V-JEPA~2~\cite{vjepa2} & Hybrid & Latent & Future latent state & Predictive prior + planning & Pre / Infer \\
      & Genie Envisioner~\cite{genie_envisioner} & VLA & Video & Action conditioned video & Predictive pretraining & Multi \\
      & PointWorld~\cite{pointworld} & VLA & 3D & Point cloud future & Geometric predictive prior & Pre \\
      & TesserAct~\cite{tesseract} & VLA & 4D & 4D scene evolution & Spatiotemporal predictive prior & Pre \\
    \midrule
    \multirow[c]{13}{1.35cm}[-0.6ex]{\centering Integrated}
      & DreamerV3~\cite{dreamerv3} & RL & Latent & Latent rollout & Imagination based policy learning & Train \\
      & Seer~\cite{seer} & IL & Video / latent & Future visual tokens & Predictive pretraining & Pre / Train \\
      & GR-1~\cite{wu2024unleashing} & VLA & Video & Future video and action & Joint world action model & Train \\
      & GR-2~\cite{gr2} & VLA & Video & Future video and action & Joint world action model & Train \\
      & WorldVLA~\cite{worldvla} & VLA & Latent & Action conditioned future & Joint world action model & Train \\
      & RynnVLA-002~\cite{rynnvla002} & VLA & Mixed & Image and action streams & Joint world action model & Train \\
      & DUST~\cite{dust} & VLA & Latent & Future and action tokens & Joint world action model & Train \\
      & CoT-VLA~\cite{cot_vla} & VLA & Video & Visual reasoning frames & Prediction augmented policy & Train \\
      & DreamVLA~\cite{dreamvla} & VLA & Mixed & Dynamic world knowledge & Prediction augmented policy & Train \\
      & VLA-JEPA~\cite{vla_jepa} & VLA & Latent & Predictive latent state & Prediction augmented policy & Train \\
      & DIAL~\cite{dial} & VLA & Latent & Intent action latent & Prediction augmented policy & Train \\
      & FlowVLA~\cite{flowvla} & VLA & Flow & Optical flow and frame & Prediction augmented policy & Train \\
      & 3D-VLA~\cite{3d_vla} & VLA & 3D & Goal image and point cloud & Prediction augmented policy & Train \\
    \midrule
    \multirow[c]{5}{1.35cm}[-0.5ex]{\centering Planner}
      & PlaNet~\cite{planet2018} & RL & Latent & Latent rollout & Trajectory planning & Infer \\
      & UniPi~\cite{unipi} & IL & Video & Future video trajectory & Trajectory planning & Infer \\
      & SuSIE~\cite{susie} & IL & Image & Subgoal image & Subgoal planning & Infer \\
      & PIN-WM~\cite{li2025pinwm} & IL & Physics & Contact dynamics & Physics informed planning & Infer \\
      & MinD~\cite{mind} & VLA & Mixed & Low resolution future latent & Visual planning + risk monitoring & Infer \\
    \midrule
    \multirow[c]{17}{1.35cm}[-0.8ex]{\centering Infrastructure}
      & DreamGen~\cite{dreamgen} & IL & Video & Synthetic robot video & Synthetic data generation & Post \\
      & GigaWorld-0~\cite{gigaworld-0} & VLA & Video & Synthetic experience & Synthetic data generation & Post \\
      & WristWorld~\cite{wristworld} & VLA & 4D & Wrist view 4D future & Synthetic data generation & Post \\
      & World4RL~\cite{world4rl} & RL & Video & Simulated rollout & Learned environment & Post \\
      & WorldGym~\cite{worldgym} & VLA & Video & Action conditioned rollout & Learned environment + scoring & Post \\
      & World-Env~\cite{world_env} & VLA & Video & Rollout and reward & Learned environment + reward & Post \\
      & WMPO~\cite{wmpo} & VLA & Video & Imagined interaction & Policy optimization in imagination & Post \\
      & NORA-1.5~\cite{nora15} & VLA & Mixed & Preference reward & Preference scoring & Post \\
      & SRPO~\cite{srpo} & VLA & Latent & Progress signal & Progress scoring & Post \\
      & AtomVLA~\cite{atomvla} & VLA & Latent & Action chunk score & Policy post training & Post \\
      & WoVR~\cite{wovr} & VLA & Mixed & Rollout reliability & Rollout verification & Post \\
      & GPC~\cite{gpc_diffusion} & IL & Video & Action conditioned rollout & Candidate filtering & Infer \\
      & WorldPlanner~\cite{worldplanner} & IL & Video & Candidate futures & Search based evaluation & Infer \\
      & VLA-Reasoner~\cite{vla_reasoner} & VLA & Mixed & Candidate futures & Search based evaluation & Infer \\
      & AdaPower~\cite{adapower} & VLA & Mixed & Adapted future state & Test time adaptation & Infer \\
      & Self-Correcting VLA~\cite{self_correcting_vla} & VLA & Mixed & Progress forecast & Online correction & Infer \\
      & World Action Verifier~\cite{wav} & VLA & Mixed & Plausibility / reachability & Feasibility verification & Infer \\
    \bottomrule
  \end{tabular*}
\end{table*}

The previous sections organized world models by what they represent, how their predictions connect to action, and how they serve as learning or decision infrastructure. A complementary question is \emph{when} the same predictive capability enters the robot-learning pipeline. The lifecycle view is useful because the role, risk, and evaluation criterion of a world model change across stages. During pretraining, a world model mainly learns a reusable predictive structure. During post-training, it becomes an operational tool for improving a policy after initial supervised or imitation learning. During inference, it serves as a mechanism for online reasoning, adaptation, and correction based on the current scene and execution history. Figure~\ref{fig:lifecycle} summarizes this stage-wise shift, and Table~\ref{tab:integrated_wm_roles} connects the lifecycle view to the representation and functional taxonomies developed above.

\subsection{Pretraining}

In pretraining, the world model functions as a predictive prior. The goal is not yet to solve a specific manipulation task, but to absorb regularities about dynamics, geometry, object motion, embodiment, and action consequences that can later support many downstream policies. Predictive pretraining is attractive because robot interaction data are expensive, sparse rewards are difficult to obtain, and future prediction supplies dense supervision from trajectories that may not contain explicit task labels.

In reinforcement learning, Dreamer-style latent dynamics models illustrate this role: learning a compact predictive state space provides the foundation for imagined control~\cite{dreamerv3}. In manipulation, the same principle increasingly appears in action-aware pretraining on large robot datasets. Seer couples visual forecasting with inverse dynamics, making future prediction directly useful for downstream action learning rather than only for representation extraction~\cite{seer}. Foundation-scale efforts push this idea further. Genie Envisioner treats large-scale video world modeling as a substrate for policy learning, evaluation, and simulation, while PointWorld and TesserAct argue that transferable predictive priors may need stronger geometric and spatiotemporal structure through point-cloud or four-dimensional scene modeling~\cite{genie_envisioner,pointworld,tesseract}.

The main advantage of pretraining is breadth. A well-pretrained world model can encode reusable dynamics across tasks, scenes, objects, and embodiments before any task-specific optimization begins. The main risk is weak alignment between actions. A model can learn visually or geometrically rich predictive features that transfer poorly to control if its pretraining objective does not preserve action consequences, contact structure, or embodiment-specific feasibility. Pretraining should therefore be evaluated not only by reconstruction or prediction metrics, but by transfer to downstream policies, robustness under new objects and viewpoints, and ablations that separate generic representation learning from genuinely action-conditioned world modeling.

\subsection{Post-Training}

Post-training begins after an initial policy has been learned, often through imitation learning, supervised VLA training, or a large-scale foundation-policy recipe. At this stage, the world model shifts from being a broad prior to being an operational improvement mechanism. It can generate additional supervision, provide a learned environment for policy refinement, supply rewards or preferences, or filter hallucinated trajectories. The bottleneck is no longer only representation learning, but how to push a policy beyond the coverage and quality of its original data without requiring large amounts of risky real-world interaction.

One post-training role is data generation. DreamGen uses a video world model to synthesize robot videos in novel environments and for novel behaviors, then recovers pseudo-actions through latent-action or inverse-dynamics models to support continued policy learning~\cite{dreamgen}. This line shows that post-training is not synonymous with reinforcement learning. A world model can improve a policy by generating new supervised experience, provided the generated trajectories remain action-consistent and embodiment-compatible.

A second role is simulation for policy refinement. World4RL uses a frozen diffusion world model as a high-fidelity simulator for manipulation policy improvement~\cite{world4rl}. VLA-oriented systems extend the same idea to generalist policies trained primarily by imitation. World-Env and VLA-RFT use learned simulators to reduce the need for risky or irreversible physical interaction during reinforcement-style refinement, while WMPO argues that pixel-space rollouts can be advantageous when they align with visual features inherited from VLA pretraining~\cite{world_env,vla_rft,wmpo}. These methods convert world models from passive predictors into training environments.

A third role is reward, preference, and progress estimation. NORA-1.5 uses world-model-informed preference rewards for DPO-style adaptation, SRPO uses world-model latent representations as a dense progress metric, and AtomVLA scores candidate action chunks against decomposed subtask goals during offline post-training~\cite{nora15,srpo,atomvla}. WoVR adds an important reliability perspective by showing that post-training in learned worlds is effective only when hallucination and rollout depth are explicitly controlled~\cite{wovr}. Across these efforts, the central post-training risk is simulator exploitation. Once a policy is optimized inside an imagined world, it may discover artifacts that improve imagined reward without improving real manipulation. Post-training should therefore be evaluated by real-world policy improvement, preservation of policy rankings, robustness to rollout depth, and explicit checks for physically invalid or unexecutable imagined transitions.

\subsection{Inference Adaptation}

At inference time, the world model operates under the strongest constraints. It must improve decisions in the current scene quickly enough to affect execution. The goal is no longer broad pretraining or offline policy improvement, but online reasoning under partial observability, distribution shift, and execution uncertainty. Inference-time world modeling can take the form of search, candidate reranking, test-time model adaptation, memory update, progress forecasting, or self-correction.

One family of methods uses predictive rollouts for test-time planning and search. VLA-Reasoner augments an off-the-shelf VLA with Monte Carlo tree search over imagined futures, while MinD couples a low-frequency visual generator with a high-frequency action model so that the policy can act on predictive latent futures and monitor risk~\cite{vla_reasoner,mind}. Another family adapts the world model itself at test time. AdaPower specializes in a pretrained world foundation model through test-time training and memory persistence, keeping predictive control aligned with the current environment~\cite{adapower}. A lighter-weight alternative is to use sparse predictive signals for online correction without full replanning. Self-Correcting VLA forecasts short-term progress and trajectory trends, then uses these predictions to refine actions during execution~\cite{self_correcting_vla}.

The main advantage of inference adaptation is specificity. The world model can condition on the exact scene, history, and task instance rather than relying only on offline training. The main risk is latency and miscalibration. A slow predictor may be unusable in closed-loop control, and a miscalibrated predictor may make the system worse by confidently selecting an imagined future that cannot be realized. Inference-time world models should therefore be evaluated by online success under distribution shift, recovery from execution errors, compute latency, and calibration between predicted and realized progress.

\subsection{Lifecycle-Level Design Implications}

The lifecycle view clarifies why a single metric cannot evaluate all world models. A pretrained world model should be judged by transfer, coverage, and action alignment. A post-training world model should be judged by whether imagined experience improves real policy performance without simulator exploitation. An inference-time world model should be judged by whether its predictions improve online decisions under latency constraints. The same architecture may appear in all three stages, but its failure modes change with use. A video predictor used for pretraining may fail to learn active visual regularities. The same predictor used for post-training may fail to generate executable supervision. Used at inference time, it may fail due to being too slow or overly confident.

This stage dependence also explains why world models are becoming infrastructure rather than isolated modules. A foundation-scale predictive model may be pretrained on broad robot data, reused as a simulator during post-training, and queried again for online correction at deployment. Each reuse changes the interface between prediction and action. The lifecycle perspective, therefore, complements the representation, prediction--action, and infrastructure taxonomies: representation determines what can be predicted, the action interface determines how prediction affects behavior, infrastructure roles determine how prediction is reused across a system, and lifecycle stage determines which risks and evaluations matter most.

\section{Datasets for World-Model Learning}
\label{sec:datasets}

World-model learning imposes different data requirements than ordinary policy learning. A policy dataset can be useful if it maps observations and instructions to successful actions. A world-model dataset must additionally support prediction: it should expose how observations, objects, contacts, and robot states evolve under intervention. The most useful datasets, therefore, contain temporally aligned observations and actions, sufficient diversity for counterfactual generalization, and modalities that reveal the physical variables relevant to manipulation. RGB video alone can support visual forecasting, but contact-rich world models benefit from depth, point clouds, force, tactile, audio, event streams, proprioception, and multi-view observations.

\begin{figure*}[!t]
\centering
\includegraphics[width=\textwidth]{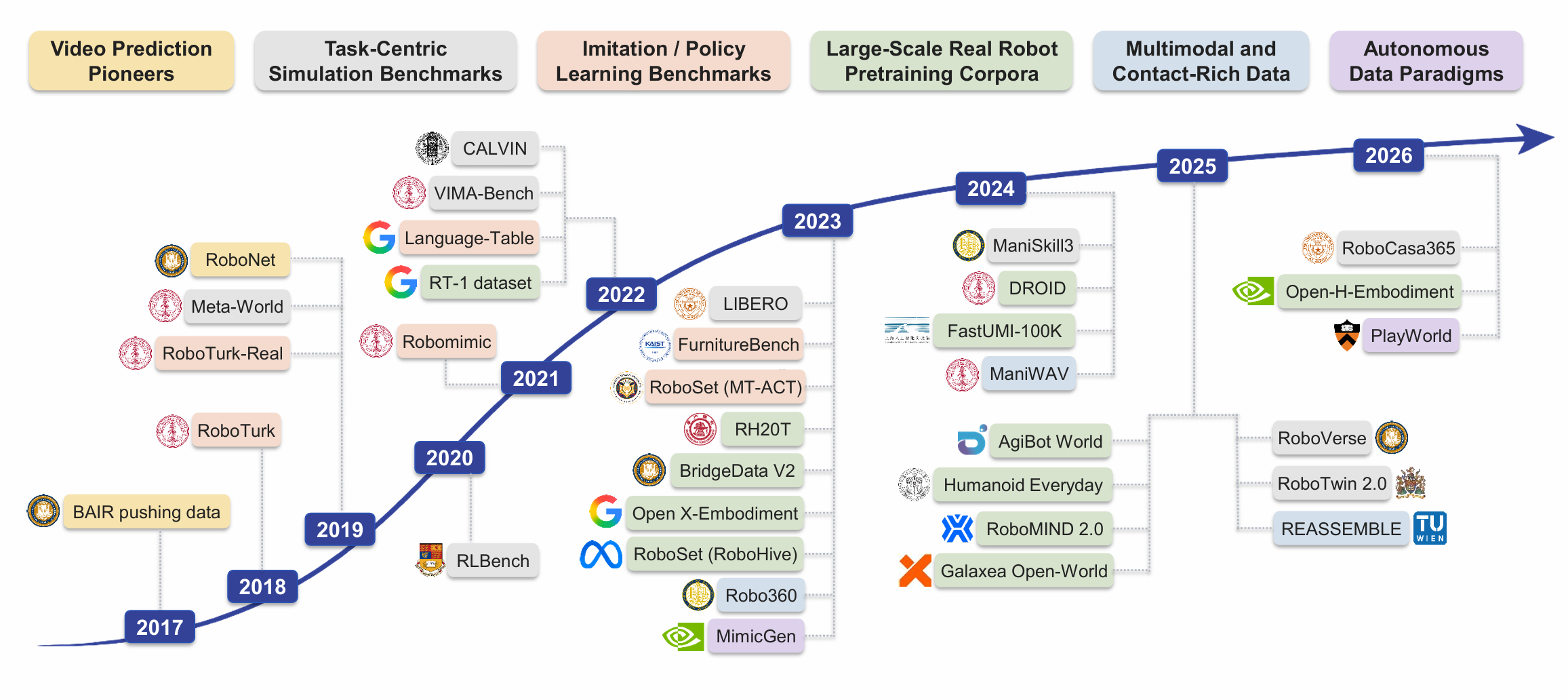}
\caption{Chronological overview of the 34 representative manipulation datasets in Table~\ref{tab:datasets}, with the horizontal axis denoting first public release year and color-coded headers matching the functional groups~A--F. The timeline highlights the shift from early video-prediction corpora to simulation benchmarks, imitation-learning resources, large-scale real-robot pretraining corpora, multimodal/contact-rich datasets, and autonomous data-collection paradigms.}
\label{fig:datasets-timeline}
\end{figure*}

Table~\ref{tab:datasets} surveys 34 principal data sources that have shaped world-model learning or its evaluation in robotic manipulation, and Fig.~\ref{fig:datasets-timeline} provides a chronological view. We include datasets and benchmarks that satisfy at least one of three criteria: they provide action-conditioned trajectories suitable for predictive modeling, they define task suites commonly used to evaluate world-model-augmented policies, or they supply multimodal, cross-embodiment, or autonomous data that addresses a known bottleneck in predictive robot learning. We organize the table by functional role rather than by modality or release date because the central question is how each resource supports the world-model pipeline. Early video-prediction corpora provided action-conditioned visual dynamics. Simulation benchmarks standardized tasks and success metrics. Demonstration collections exposed how data quality and collection strategy affect downstream policies. Large-scale real-robot corpora enabled predictive pretraining. Multimodal datasets added signals for contact and physical interaction. Autonomous data paradigms reduced dependence on human teleoperation.

Groups~B and~C both support imitation learning but differ in emphasis. Group~B defines canonical task suites and evaluation protocols. Group~C focuses on how demonstrations are collected, structured, and reused. This distinction matters for world models because evaluation benchmarks test whether prediction improves behavior, whereas demonstration datasets determine what dynamics and failure modes the model can learn.

\begin{table*}[!t]
\centering
\caption{Representative manipulation datasets categorized by functional role in the survey taxonomy. Each row summarizes a dataset's release year, domain, scale, task coverage, embodiment diversity, sensing modalities, and primary functional role in the world-model pipeline. Categories are not mutually exclusive; the table reports the role most emphasized by each dataset.}
\label{tab:datasets}
\scriptsize
\setlength{\tabcolsep}{2.6pt}
\begin{tabular}{@{}llccllll@{}}
\toprule
\textbf{Dataset} & \textbf{Year} & \textbf{Dom.} & \textbf{Scale} & \textbf{Tasks} & \textbf{Embodiment} & \textbf{Mod.} & \textbf{Functional Role} \\
\midrule
\multicolumn{8}{@{}l}{\textit{A. Video Prediction Pioneers}} \\
BAIR pushing data~\cite{bair_pushing} & 2017 & Real & 44K traj. & --$^{\mathrm{s}}$ & 1 (Sawyer) & R & Video pred. \\
RoboNet~\cite{robonet} & 2019 & Real & ${\sim}$162K traj. & --$^{\mathrm{s}}$ & 7 robots & R, P & Cross-embod. pretrain \\
\midrule
\multicolumn{8}{@{}l}{\textit{B. Task-Centric Simulation Benchmarks}} \\
Meta-World~\cite{metaworld}$^{\mathrm{m}}$ & 2019 & Sim & --$^{\mathrm{s}}$ & 50 & 1 (Sawyer) & state/P & RL bench. \\
RLBench~\cite{rlbench}$^{\mathrm{m}}$ & 2020 & Sim & infinite demos & 100 & 1 (Franka) & R, D, Seg, P, L & IL/RL bench. \\
CALVIN~\cite{calvin}$^{\mathrm{m}}$ & 2022 & Sim & ${\sim}$24 h play & 34 & 1 (Franka) & R, D, P, L, T & IL/VLA bench. \\
VIMA-Bench~\cite{vima_bench}$^{\mathrm{m}}$ & 2022 & Sim & 650K traj. & 17 task templates & 1 (UR5) & R,L,Seg (part.) & IL/VLA bench., foundation \\
LIBERO~\cite{libero}$^{\mathrm{m}}$ & 2023 & Sim & 50 demos/task & 130 (4 suites) & 1 & R, P, L & Lifelong IL bench. \\
ManiSkill3~\cite{maniskill3}$^{\mathrm{m}}$ & 2024 & Sim & M{+} demo fr. & 12 domains & 20{+} robots & task-dep.$^{\mathrm{a}}$ & RL, data gen.$^{\mathrm{e}}$ \\
RoboVerse~\cite{roboverse}$^{\mathrm{m}}$ & 2025 & Sim & 510.5K traj. & 276 cat. & Multi & src.-dep. & Pretrain, bench. \\
RoboTwin 2.0~\cite{robotwin}$^{\mathrm{m}}$ & 2025 & Sim & 100K{+} traj. & 50 & 5 dual-arm & R, P, L & Bench., data gen. \\
RoboCasa365~\cite{robocasa365}$^{\mathrm{m}}$ & 2026 & Sim & ${\sim}$655K demos / 2227 h$^{\mathrm{k}}$ & 300$^{\mathrm{k}}$ & 1 mobile manip. & R, P, L & Pretrain, bench. \\
\midrule
\multicolumn{8}{@{}l}{\textit{C. Imitation / Policy Learning Benchmarks}} \\
RoboTurk~\cite{roboturk} & 2018 & Sim & 2.2K demos / 137.5 h & 2 / 3$^{\mathrm{r}}$ & 1 (Sawyer) & R; P part. & Crowdsourced IL$^{\mathrm{l}}$ \\
RoboTurk-Real~\cite{roboturk_real} & 2019 & Real & 2.1K demos / 111 h & 3 & 1 (Sawyer) & R, D, P & Crowdsourced IL \\
Robomimic~\cite{robomimic}$^{\mathrm{m}}$ & 2021 & R+S & ${\sim}$2.8K demos + MG$^{\mathrm{d}}$ & 8 & Franka Panda$^{\mathrm{q}}$ & R, P & Offline IL bench. \\
Language-Table~\cite{language_table}$^{\mathrm{m}}$ & 2022 & R+S & ${\sim}$600K traj. & 5 fam.; 696 cond. & 1 (xArm) & R, L & Lang.-cond. IL \\
FurnitureBench~\cite{furniturebench}$^{\mathrm{m}}$ & 2023 & R+S & 219.6 h / 5100 demos & 8+one\_leg$^{\mathrm{p}}$ & 1 (Franka) & R, D, P & IL/RL bench. \\
RoboSet (MT-ACT)~\cite{roboagent} & 2023 & Real & 7.5K traj. & 38 & 1 (Franka) & R, D, P, L & Semantic aug. IL \\
\midrule
\multicolumn{8}{@{}l}{\textit{D. Large-Scale Real-Robot Pretraining Corpora}} \\
RT-1 dataset~\cite{rt1_dataset} & 2022 & Real & ${\sim}$130K ep. & 700{+} instr. & 1 (ER)$^{\mathrm{b}}$ & R, L & Foundation, IL bench.$^{\mathrm{f}}$ \\
RH20T~\cite{rh20t} & 2023 & Real & 110K seq. & 147 & 4 arms / 7 configs & R, D, F, A, P, L; part. T & Pretrain \\
BridgeData V2~\cite{bridgev2} & 2023 & Real & ${\sim}$60K traj. & 13 skills & 1 (WidowX) & R,L; part. D & Pretrain \\
Open X-Embodiment~\cite{oxe} & 2023 & Real & 1M{+} traj. & 160K{+} labels$^{\mathrm{g}}$ & 22 & src.-dep. & Pretrain \\
RoboSet (RoboHive)~\cite{robohive} & 2023 & R+S & 99.6K traj.$^{\mathrm{h}}$ & 64$^{\mathrm{h}}$ & Multi (Franka, etc.) & R, P; D src.-dep. & Unified sim+real \\
DROID~\cite{droid} & 2024 & Real & 76K traj. & 86 (verbs)$^{\mathrm{i}}$ & 1 (Franka) & R, D, P, L & Pretrain \\
FastUMI-100K~\cite{fastumi} & 2024 & Real & 100K{+} traj. / 600 h & 54 & handheld$^{\mathrm{o}}$ & R, P, L & IL/VLA bench., pretrain \\
AgiBot World~\cite{agibot_world} & 2025 & Real & 1M{+} traj. & 217 & 1 (AgiBot G1) & R,D,P,L; part. T & Foundation \\
Galaxea Open-World~\cite{galaxea_openworld} & 2025 & Real & 500 h / 100K traj. & 150 & 1 (R1 Lite) & R,P,L; part. D & VLA bench., pretrain \\
Humanoid Everyday~\cite{humanoid_everyday} & 2025 & Real & 10.3K traj. & 260 & 2 humanoids & R, D, P, Li, L; part. T & Humanoid pretrain \\
RoboMIND 2.0~\cite{robomind2} & 2025 & R+S & 330K traj.$^{\mathrm{j}}$ & 759$^{\mathrm{j}}$ & 6 dual-arm & R,D,P,L,F; part. T & Pretrain, bench. \\
Open-H-Embodiment~\cite{ohe} & 2026 & R+S & 124K ep. / 770 h & 33 fam.$^{\mathrm{n}}$ & 20 & R,P; part. D, L, Seg, F & Medical pretrain \\
\midrule
\multicolumn{8}{@{}l}{\textit{E. Multimodal and Contact-Rich Data}} \\
Robo360~\cite{robo360} & 2023 & Real & 2K{+} traj. & 5 rep. tasks$^{\mathrm{t}}$ & 1 (xArm6) & R${\times}$86,D,3D,P,A & 3D/4D WM \\
ManiWAV~\cite{maniwav} & 2024 & Real & 557/119/145/193 demos & 4 & 1 (UR5) & R, A, P & Audio-visual IL \\
REASSEMBLE~\cite{reassemble} & 2025 & Real & 4.6K demos / 781 min & 68 (+Idle) & 1 (Franka FR3) & R, Ev, P, L, F, A & Multimodal assembly \\
\midrule
\multicolumn{8}{@{}l}{\textit{F. Autonomous Data Paradigms}} \\
MimicGen~\cite{mimicgen} & 2023 & R+S & 50K{+} gen.\ demos & 18 & 4 sim arms$^{\mathrm{c}}$ & R, P & Data gen., IL bench. \\
PlayWorld~\cite{playworld} & 2026 & Real & 30 h auton. play & --$^{\mathrm{s}}$ & 1 (DROID setup) & R,P; part. L & WM data engine \\
\bottomrule
\end{tabular}
\par\smallskip
\noindent
\begin{minipage}{\linewidth}
\scriptsize
\textit{Notes.} Datasets are ordered chronologically by first public release year within each functional group. The \textbf{Year} column reports the first public release year. The \textbf{Tasks} column is not directly comparable across datasets because entries may denote benchmark tasks, skills, instruction labels, templates, domains, or families; ``--'' marks information that is not reported or not applicable in the corresponding column (see note~s). \textbf{Modality codes}: R=RGB, D=Depth, 3D=3D vision, Ev=event camera, P=Proprioception, state=low-dimensional task/object/goal state, L=Language, Seg=Segmentation, F=Force/Torque, A=Audio, T=Tactile, Li=LiDAR; action labels/control commands are implicit for trajectory datasets and omitted from \textbf{Mod.}; src.-dep.\ = source-dependent; part.\ = partial. R+S=Real+Sim.
\par\smallskip
\begin{tabular*}{\linewidth}{@{\extracolsep{\fill}}p{0.48\linewidth}p{0.48\linewidth}@{}}
\raggedright $^{\mathrm{a}}$\,ManiSkill3: task/framework-dependent modalities; R/D/3D/P/L/Seg/T possible.\par
$^{\mathrm{b}}$\,ER=Everyday Robots; Kuka only in external-gen. tests.\par
$^{\mathrm{c}}$\,MimicGen also reports real-world validation.\par
$^{\mathrm{d}}$\,MG=machine-generated; Robomimic also includes MG transition data.\par
$^{\mathrm{e}}$\,ManiSkill3 used as RL and data-generation benchmark.\par
$^{\mathrm{f}}$\,RT-1 dataset also used as foundation-model evaluation benchmark.\par
$^{\mathrm{g}}$\,OXE: 160{,}266 instruction/task labels, not benchmark tasks.\par
$^{\mathrm{h}}$\, RoboHive Table\,2 aggregates 99{,}550 trajectories and 64 tasks across real kitchen, bin-manipulation, and 775 simulation demos; this is a broader sim+real counting scheme than RoboAgent's real-world RoboSet full (98{,}050 trajectories, 48 tasks).\par
$^{\mathrm{i}}$\,DROID counts de-duplicated instruction verbs as tasks.\par
$^{\mathrm{j}}$\,RoboMIND 2.0: 310K real + 20K sim.; Table\,1 lists 759 tasks, abstract 739.\par
$^{\mathrm{k}}$\,RoboCasa365: 55K human + 600K synthetic demos (2227 h); 365 benchmark tasks; 300 pretraining tasks; 50 target eval tasks.
&
\raggedright $^{\mathrm{l}}$\,RoboTurk: video/action present; proprioceptive fields not fully enumerated.\par
$^{\mathrm{m}}$\,Also used as evaluation benchmark.\par
$^{\mathrm{n}}$\,Open-H-Embodiment: medical/surgical domain, aggregating data from 49{+} institutions.\par
$^{\mathrm{o}}$\,FastUMI-100K: handheld gripper data collection; evaluated on two robot platforms.\par
$^{\mathrm{p}}$\, FurnitureBench: 8 furniture models plus one\_leg reproducibility/single-leg task in the reported statistics.\par
$^{\mathrm{q}}$\, Robomimic includes both single-arm and dual-arm Franka Panda setups.\par
$^{\mathrm{r}}$\, RoboTurk collected demonstrations for 2 tasks and evaluated transfer on 3 tasks.\par
$^{\mathrm{s}}$\, ``--'' marks information not reported by the source or not applicable to that column.\par
$^{\mathrm{t}}$\,Robo360 reports five representative policy-learning tasks: towel-to-basket, organize slippers, flip bottle upright, cable-to-basket, and rope-to-basket, within a broader multi-material corpus.
\end{tabular*}
\end{minipage}
\end{table*}

\subsection{Video Prediction Pioneers}

The earliest manipulation datasets for world modeling were built around action-conditioned visual prediction. BAIR pushing data~\cite{bair_pushing} established a simple but influential setting: a robot executes random pushing actions, and the model predicts how the image changes. Its value was not task diversity, but clean temporal supervision for learning visual dynamics. RoboNet~\cite{robonet} extended this idea to seven robots across multiple institutions, providing synchronized RGB video, robot states, and actions across approximately 162K trajectories. It showed that visual dynamics models trained on diverse embodiments can transfer to unseen platforms with modest fine-tuning, anticipating the cross-embodiment pretraining philosophy later pursued by larger robot-data initiatives.

These datasets also reveal an early limitation. Random interaction produces broad local dynamics but weak task structure. It teaches a model how objects move under robot actions, but not necessarily which futures are useful for manipulation. Later datasets, therefore, expanded along two directions: more structured tasks for evaluation, and larger, more diverse trajectories for pretraining.

\subsection{Task-Centric Simulation Benchmarks}

Simulation benchmarks address a different need: standardized evaluation. They define tasks, initial conditions, observations, and success metrics that enable comparisons of policies and planners under controlled conditions. Meta-World~\cite{metaworld} defined 50 parametric tasks for multi-task and meta-RL, while RLBench~\cite{rlbench} broadened coverage to 100 tasks with richer observations and procedural demonstration generation. CALVIN~\cite{calvin} introduced long-horizon language-conditioned evaluation through chained subgoals, VIMA-Bench~\cite{vima_bench} formalized multimodal prompting and compositional generalization, and LIBERO~\cite{libero} decomposed transfer into spatial, object, goal, and compositional suites.

For world models, these benchmarks are useful because they test whether prediction improves downstream behavior rather than only one-step reconstruction. A latent dynamics model can be judged by sample efficiency on Meta-World or RLBench. A VLA policy with predictive reasoning can be evaluated on CALVIN or LIBERO through long-horizon success. A simulator-like world model can be stress-tested by whether policies improved in imagination transfer back to benchmark tasks. The limitation is that many simulated suites simplify the very dynamics that make manipulation difficult. Contacts, deformation, friction, tool use, and rare failures are often less diverse than in real scenes, which limits their ability to validate world models intended for physical deployment.

Recent simulation resources broaden the scale and realism of this evaluation layer. ManiSkill3~\cite{maniskill3} emphasizes GPU-parallelized throughput across many embodiments, making RL-scale data generation more practical. RoboVerse~\cite{roboverse} unifies multiple simulators under a common interface, reducing friction in cross-platform evaluation. RoboTwin~2.0~\cite{robotwin} expands dual-arm coverage, and RoboCasa365~\cite{robocasa365} pushes toward household-scale pretraining and evaluation. These benchmarks make it easier to test world models across tasks and embodiments, but they do not remove the need for real-world validation.

\subsection{Demonstration Collection and Imitation Learning}

Demonstration datasets reveal how the source and quality of trajectories shape both policy learning and predictive modeling. RoboTurk~\cite{roboturk} pioneered crowdsourced teleoperation through smartphone-based 6-DoF control, and RoboTurk-Real~\cite{roboturk_real} extended this paradigm to physical robots. These datasets showed that demonstration quality, strategy diversity, and operator variability matter as much as raw scale. Robomimic~\cite{robomimic} made this point systematic by varying demonstrator proficiency, dataset size, and observation modality across several manipulation tasks.

For world models, demonstration data have a dual role. They provide action-conditioned trajectories for learning dynamics and encode the distribution of purposeful behavior that futures generated should respect. This is different from random play. A world model trained only on successful demonstrations may learn task-relevant transitions but lack failure and recovery dynamics. A model trained on broader teleoperation data may capture more diverse states but require stronger filtering to distinguish useful futures from irrelevant behavior. Language-Table~\cite{language_table}, RoboSet (MT-ACT)~\cite{roboagent}, and FurnitureBench~\cite{furniturebench} extend this line by adding language grounding, semantic augmentation, or long-horizon assembly structure. Their value for world modeling lies not only in action labels but in connecting predictive dynamics to goals, instructions, and compositional task progress.

\subsection{Large-Scale Real-Robot Pretraining Corpora}

Large-scale real-robot corpora changed the role of datasets from task-specific supervision to foundation-model pretraining. RT-1~\cite{rt1_dataset} initiated this scaling trend with approximately 130K episodes across more than 700 instructions, showing that task diversity can matter more than repeated demonstrations of a narrow skill. Open X-Embodiment~\cite{oxe} aggregated over one million trajectories across many robots and institutions, validating cross-embodiment pretraining while exposing the difficulty of heterogeneous action spaces, camera setups, and annotation conventions. DROID~\cite{droid} took a complementary approach by standardizing the embodiment and hardware while varying environments across many locations. AgiBot World~\cite{agibot_world} pushed homogeneous collection to an industrial scale with more than one million trajectories from many identical robots.

These corpora are central to modern world models because predictive pretraining depends on breadth. A world model intended to support manipulation across tasks must see many objects, backgrounds, embodiments, and action consequences. BridgeData~V2~\cite{bridgev2} emphasizes environmental breadth, RH20T~\cite{rh20t} adds richer sensing including force, audio, and partial tactile signals, and RoboHive~\cite{robohive} provides a unified sim-and-real framework. RoboMIND~2.0~\cite{robomind2}, Galaxea Open-World~\cite{galaxea_openworld}, Humanoid Everyday~\cite{humanoid_everyday}, and Open-H-Embodiment~\cite{ohe} further extend the landscape toward dual-arm, humanoid, open-world, and surgical manipulation.

The main unresolved issue is heterogeneity. Large aggregated corpora provide scale but mix embodiments, control frequencies, action parameterizations, camera viewpoints, language conventions, and task definitions. This diversity is valuable for representation learning, but it complicates action-conditioned prediction. A world model must either normalize these differences, learn embodiment-conditioned dynamics, or restrict transfer to subsets with compatible sensing and control. Scale alone is therefore insufficient. The useful question is whether the corpus exposes the variations needed for robust prediction while preserving enough structure for action alignment.

\subsection{Multimodal and Contact-Rich Data}

Most manipulation world models remain visually dominated, yet many manipulation failures stem from physical variables that RGB cannot reveal. Force closure, slip, friction, deformation, insertion pressure, vibration, and contact onset are often invisible or ambiguous in images. Multimodal datasets address this sensing gap by adding signals that make physical interaction more observable.

Robo360~\cite{robo360} provides dense multi-camera capture for reconstruction-based 3D and 4D world modeling. ManiWAV~\cite{maniwav} shows that audio from a contact microphone can improve policies on tasks such as flipping and pouring. REASSEMBLE~\cite{reassemble} combines RGB, event cameras, force/torque, audio, proprioception, and language for assembly, making it one of the richest resources for contact-intensive prediction. RH20T~\cite{rh20t}, although grouped with large-scale pretraining corpora, also contributes force, audio, and partial tactile streams.

These datasets point toward a necessary expansion of world modeling beyond vision. A model that predicts images may miss the contact events that determine whether a grasp will hold or an insertion will succeed. Multimodal prediction can make these events observable, but it introduces new challenges in synchronization, sensor calibration, missing modalities, and cross-platform transfer. The field still lacks widely adopted benchmarks that test whether non-visual predictive signals improve world-model reliability rather than only policy performance on isolated tasks.

\subsection{Autonomous Data Paradigms}

Human teleoperation remains expensive and biased toward the operator's skill repertoire. Autonomous and semi-autonomous data paradigms attempt to expand coverage without requiring proportional human effort. MimicGen~\cite{mimicgen} synthesizes new demonstrations from a small number of source trajectories through object-centric spatial transformations, preserving useful task structure while increasing diversity. PlayWorld~\cite{playworld} uses autonomous robot play to collect interaction data for world-model learning, suggesting that exploration diversity can compensate partly for the lack of task-directed demonstrations.

The two paradigms address different data bottlenecks. Demonstration synthesis amplifies human expertise but remains anchored to the source demonstrations. Autonomous play explores broader state-action distributions and can expose failures, contacts, and recovery behaviors that are rare in curated datasets, but it may lack semantic task structure. For world models, these paradigms are complementary. A predictive model needs both purposeful behavior and out-of-distribution interactions if it is to support planning, verification, and recovery during deployment.

\subsection{Discussion and Open Problems}

The dataset landscape reveals a supply chain for world-model learning. Early robot-video corpora provide action-conditioned visual dynamics. Simulation benchmarks provide standardized task evaluation. Demonstration datasets provide goal-directed behavior. Large real-robot corpora provide scale. Multimodal datasets expose hidden physical variables. Autonomous data paradigms broaden coverage beyond human demonstrations. Recent world models such as IRASim~\cite{irasim} already draw from upstream corpora including RT-1, BridgeData~V2, Language-Table, and RoboNet, indicating a shift from monolithic dataset collection toward modular data composition.

Four open problems remain. First, task granularity is inconsistent. A task may refer to a benchmark goal, language instruction, verb label, skill family, or generated scenario. This makes it difficult to compare coverage across datasets and to evaluate whether a world model has learned reusable dynamics or merely memorized task templates. Second, the heterogeneity of action and embodiment remains unresolved. Cross-embodiment corpora are essential for scale, but world models must still predict consequences under a specific robot's morphology, controller, and action space. Third, non-visual modalities are underused. Force, tactile, audio, event, and proprioceptive signals are increasingly available, but most world models still train primarily on RGB or RGB-D. Fourth, rare failures and contact-rich transitions remain undersampled. Million-trajectory datasets may still contain too few insertions, slips, jams, deformable interactions, or recovery behaviors to train reliable predictive models.

These limitations directly affect evaluation. A benchmark can only test the failures that its data and task protocols expose. If datasets lack contact sensing, the evaluation will understate physical inconsistency. If they contain only successful demonstrations, learned worlds may fail during recovery. If task labels are incomparable, progress metrics cannot transfer across benchmarks. The next section, therefore, turns from data resources to evaluation protocols, asking how world models should be judged when their predictions are used for control, simulation, post-training, and verification.

\section{Benchmarks and Evaluation}
\label{sec:benchmarks}

Evaluation is the central unresolved problem for world models in robotic manipulation. A predictive model can be visually accurate yet physically wrong, physically plausible yet useless for action, or useful for a policy only because it improves representation learning rather than because its rollouts are reliable. Conversely, a compact latent model may produce no interpretable image at all while still improving sample efficiency or long-horizon success. Evaluation must therefore ask not only whether the model predicts the future, but whether the predicted future remains actionable, physically credible, and useful for the policy, planner, simulator, or verifier that consumes it.

This section complements the dataset overview in Section~\ref{sec:datasets} by distinguishing three evaluation tiers: direct predictive fidelity, downstream task performance, and simulator reliability. Direct fidelity asks whether predicted observations, latents, geometry, or physical states match held-out trajectories. Downstream performance asks whether using the world model improves control, generalization, or sample efficiency. Simulator reliability asks whether imagined interaction can be trusted for policy ranking, policy improvement, reward estimation, or verification. Table~\ref{tab:metric_summary} summarizes common metrics associated with these tiers. The main lesson is that no single tier is sufficient. A world model used as a learned environment requires reliability tests that a passive video predictor may not, while a latent planner may need downstream and calibration metrics more than pixel-level reconstruction scores.

\begin{table*}[htbp]
\centering
\caption{Common metrics for evaluating world models in robotic manipulation. Direct predictive-fidelity metrics compare predicted observations or latent states with target trajectories. Downstream metrics measure the behavior of the policy or system that uses the world model. Simulator-reliability metrics test whether imagined rollouts remain physically plausible and useful for policy evaluation or improvement.}
\label{tab:metric_summary}
\footnotesize
\renewcommand{\arraystretch}{1.18}
\setlength{\tabcolsep}{3pt}
\begin{tabular}{@{}p{0.19\textwidth} >{\centering\arraybackslash}m{0.32\textwidth} p{0.41\textwidth}@{}}
\toprule
\textbf{Evaluation tier} & \textbf{Metric and formula} & \textbf{Interpretation} \\
\midrule
\multirow{4}{=}{Direct predictive fidelity}
& $\displaystyle \mathrm{PSNR}=10\log_{10}\frac{\mathrm{MAX}^{2}}{\mathrm{MSE}}$
& Pixel-level reconstruction quality. $\mathrm{MAX}$ denotes the maximum pixel value, and MSE denotes the mean squared image error. Higher is better. \\
\cmidrule{2-3}
& $\displaystyle \mathrm{SSIM}(x,y)=\frac{(2\mu_x\mu_y+c_1)(2\sigma_{xy}+c_2)}{(\mu_x^{2}+\mu_y^{2}+c_1)(\sigma_x^{2}+\sigma_y^{2}+c_2)}$
& Structural image similarity between prediction $x$ and target $y$, based on luminance, contrast, and covariance terms. Higher is better. \\
\cmidrule{2-3}
& $\displaystyle \mathrm{LPIPS}(x,y)=\sum_l w_l\, d\bigl(\phi_l(x),\phi_l(y)\bigr)$
& Learned perceptual distance between deep features $\phi_l$ of prediction and target, with layer weights $w_l$. Lower is better. \\
\cmidrule{2-3}
& $\displaystyle \mathrm{LC}=D_{\mathrm{KL}}(P\|Q)=\sum_i P(i)\log\frac{P(i)}{Q(i)}$
& Latent consistency between a predicted latent distribution $P$ and a target prior or posterior $Q$. Lower divergence indicates a more regular latent rollout. \\
\midrule
\multirow{3}{=}{Downstream task performance}
& $\displaystyle \mathrm{SR}=\frac{\text{successful episodes}}{\text{total episodes}}$
& Task success rate under a fixed evaluation protocol. Higher is better. \\
\cmidrule{2-3}
& $\displaystyle \mathrm{SE}=\inf\{\,N \mid \mathrm{SR}_N \ge \tau\,\}$
& Sample efficiency, measured as the smallest number of episodes or environment steps $N$ needed to reach target success rate $\tau$. Lower is better. \\
\cmidrule{2-3}
& $\displaystyle \mathrm{ACS}=\frac{1}{N}\sum_{i=1}^{N}\mathrm{Subtasks}_i$
& Average completed subtasks in long-horizon evaluation, where $\mathrm{Subtasks}_i$ counts consecutive subgoals completed before failure in episode $i$. Higher is better. \\
\midrule
\multirow{3}{=}{Simulator reliability}
& $\displaystyle \rho = 1 - \frac{6\sum_i d_i^2}{n(n^2-1)}$
& Spearman policy-rank correlation between simulated and real policy rankings, where $d_i$ is the rank difference for policy $i$. Higher is better. \\
\cmidrule{2-3}
& $\displaystyle \mathrm{HR}=\frac{\text{infeasible transitions}}{\text{total transitions}}$
& Hallucination rate, measured as the fraction of generated transitions that violate physical or task constraints. Lower is better. \\
\cmidrule{2-3}
& $\displaystyle \mathcal{D}=\frac{1}{T}\sum_{t=1}^{T}\|\mathbf{s}_t-\hat{\mathbf{s}}_t\|_2^2$
& Rollout divergence between real states $\mathbf{s}_t$ and predicted states $\hat{\mathbf{s}}_t$ over horizon $T$. Lower is better. \\
\bottomrule
\end{tabular}
\end{table*}

\subsection{Predictive Fidelity Metrics}
\label{subsec:direct_eval}

The most direct evaluation treats a world model as an action-conditioned predictor and compares its outputs with held-out trajectories. For image and video models, common metrics include PSNR, SSIM, LPIPS, and FVD. BAIR Robot Pushing, KITTI, and RoboNet are often used in this tradition~\cite{robonet}. These metrics are easy to compute and useful for detecting severe visual prediction errors. They also provide continuity with the video-prediction literature.

Their limitation is that they measure appearance more than manipulation relevance. PSNR and SSIM can favor blurred averages over sharp but uncertain futures. LPIPS and FVD can reward perceptual realism without testing whether contacts, object permanence, kinematic constraints, or action effects are correct. A predicted video may look plausible while showing an impossible grasp or a sliding object that violates contact dynamics. For manipulation, fidelity metrics should therefore be interpreted as necessary but insufficient diagnostics.

Recent work extends fidelity evaluation toward geometry, motion, and latent consistency. Robo360~\cite{robo360} evaluates multi-view consistency, which exposes errors hidden in single-view RGB prediction. TesserAct~\cite{tesseract} and PointWorld~\cite{pointworld} evaluate predictions through 3D point clouds, four-dimensional scene structure, and scene flow, which better capture spatial drift and object motion. Latent world models are commonly assessed through reconstruction losses, likelihood terms, KL penalties, or latent consistency objectives, as in Dreamer-style systems~\cite{dreamerv3}. These quantities are harder to compare across methods because each model defines its own latent space. The practical rule is that fidelity metrics should align with the world model's representation and intended use. Pixel metrics are appropriate for visual simulators, geometric metrics for spatial predictors, flow metrics for motion-centric models, and latent regularity for compact planning substrates.

The missing fidelity metric is action alignment. A manipulation world model should not only predict a plausible next state, but also predict the consequences of the specified robot action. This requires metrics that compare predicted and realized object displacement, contact onset, affordance change, progress toward goals, and executability under an inverse dynamics or controller. Such metrics remain less standardized than image similarity, which is one reason world-model evaluation remains fragmented.

\subsection{Downstream Task and System Benchmarks}
\label{subsec:downstream}

For robotic manipulation, the strongest evidence for a world model is downstream improvement under a fixed evaluation protocol. A useful world model should improve policy learning, planning, generalization, sample efficiency, or recovery compared with a reactive or non-predictive baseline. Simulation suites provide the most common testbeds. Meta-World~\cite{metaworld} and RLBench~\cite{rlbench} support sample-efficiency and final-success comparisons for model-based reinforcement learning. Dreamer-style methods and TD-MPC variants are often judged by the number of environment steps needed to reach a target success rate~\cite{dreamerv3,hansen2024tdmpc2}.

Long-horizon imitation and VLA benchmarks test a different failure mode. CALVIN~\cite{calvin} and LIBERO~\cite{libero} require language-conditioned systems to execute sequences of subtasks. Here, the world model is useful only if it helps maintain progress over multiple decisions, avoid compounding errors, and recover from partial failures. The average completed subtasks per episode is therefore more informative than one-step prediction accuracy. Generalist benchmarks such as RoboVerse~\cite{roboverse}, ManiSkill3~\cite{maniskill3}, and RoboCasa365~\cite{robocasa365} broaden evaluation across objects, scenes, and embodiments. Held-out BridgeData V2 splits~\cite{bridgev2} and unseen instances in AgiBot World~\cite{agibot_world} further test whether predictive structure supports transfer rather than only fitting the training distribution.

The central difficulty in downstream evaluation is attribution. If a world-model-augmented policy outperforms a baseline, the gain may come from better representations, additional imagined data, auxiliary prediction losses, test-time search, reward shaping, or a combination of these factors. Fast-WAM~\cite{fast_wam} illustrates this issue by suggesting that predictive representation learning can explain much of the benefit in some systems, even when the architecture appears to support explicit test-time imagination. Strong downstream evaluation should therefore include ablations that separate world-model pretraining, rollout use, search, synthetic data, reward scoring, and policy capacity. Without such ablations, task success alone cannot reveal how prediction helped.

\subsection{Infrastructure and Simulator Reliability}
\label{subsec:infra_eval}

When world models are used as infrastructure, evaluation must go beyond both visual fidelity and task success. A learned simulator may be visually convincing and still rank policies incorrectly. A reward-producing world model may assign high scores to physically impossible trajectories. A verifier may fail to detect the same hallucinations that the generator produces. Reliability evaluation assesses whether imagined interaction remains trustworthy when used for policy selection, post-training, or closed-loop optimization.

WorldGym~\cite{worldgym} proposes evaluating the world model as a gym-like environment through policy-rank correlation. A reliable simulator should rank policy checkpoints in the same order as real-world execution. If a policy variant improves on the robot, it should also improve in the learned environment. A mismatch indicates that the simulator is not yet a dependable proxy for policy selection. This criterion is especially important for post-training, where policy optimization may otherwise exploit model artifacts.

Reliability evaluation also needs to measure hallucination and rollout drift. WoVR~\cite{wovr} and World-VLA-Loop~\cite{world-vla-loop} emphasize these failure modes. Hallucination rate measures how often the world model predicts infeasible transitions, such as object penetration, teleportation, disappearing objects, or impossible contacts. Rollout divergence measures how quickly imagined trajectories depart from realistic state sequences. These errors matter especially in contact-rich manipulation, because small geometric or force errors can change the outcome of grasping, insertion, assembly, or tool use. WMPO~\cite{wmpo} adds a deployment-oriented criterion: a learned environment should support policy improvement without extensive hyperparameter retuning or domain randomization.

The most stringent reliability test is closed-loop exploitation. A policy trained or searched within a world model actively optimizes against its own predictions. This can reveal errors that passive validation misses. A simulator-centric world model should therefore be evaluated under adversarial or optimization-heavy use: can a policy exploit visual artifacts, reward loopholes, contact hallucinations, or termination mistakes? If so, the model may still be useful for short-horizon ranking or data augmentation, but unsafe as an autonomous post-training environment.

\subsection{Open Challenges in Benchmarking}
\label{subsec:open_challenges}

Despite progress, benchmark design still lags behind how world models are used today. The first gap is contact-rich evaluation. Most standard suites emphasize rigid-body pick-and-place, reaching, pushing, or rearrangement, while cable routing, cloth folding, peg insertion, assembly, tool use, and deformable-object manipulation require metrics that capture forces, constraints, slip, jamming, and physical plausibility. Vision-only prediction scores cannot detect many of these failures.

The second gap is metric fragmentation. No widely adopted metric jointly captures visual quality, physical correctness, action alignment, downstream utility, and simulator reliability. As a result, papers often report the metric most favorable to their system class. Generative models emphasize FVD, LPIPS, or visual quality. Latent planners emphasize task success or sample efficiency. Learned simulators emphasize policy improvement or policy ranking. Verification systems emphasize hallucination or reachability. These metrics are all useful, but they are not interchangeable.

The third gap is action-conditioned calibration. A manipulation world model should know when its prediction under a candidate action is unreliable. This is essential for planning, filtering, searching, and learning environment post-training. Yet few benchmarks test calibration between predicted and realized progress, or between model confidence and action feasibility. Calibration is especially important under distribution shift, where the most dangerous predictions are often the most confident hallucinations.

The fourth gap is closed-loop exploitation. When policies are trained or searched inside a learned world, they may discover simulator-specific shortcuts rather than transferable manipulation skills. Future benchmarks should include held-out physical test scenes, adversarial perturbations, explicit infeasible-transition checks, rollout-depth stress tests, and policy-optimization protocols that assess whether imagined improvements transfer to the real robot. These protocols should stress-test the world model and policy as a coupled system rather than as separate modules.

These gaps share a common cause: many current metrics were inherited from adjacent fields rather than designed for world models. PSNR and SSIM came from image reconstruction, where pixel fidelity is the objective. Success-rate suites were built to compare policies under fixed observation and action spaces. Simulator benchmarks were designed around analytic engines rather than learned, failure-prone predictors. World-model-native evaluation should therefore measure not only whether predictions look correct, but whether they remain action-aligned, physically credible, calibrated, and robust under closed-loop policy use.

\section{Conclusion}

World models for robotic manipulation are evolving from task-specific dynamics predictors into general predictive infrastructure for robot learning. This survey has argued that this shift is best understood through three linked questions: what future representation the model predicts, how prediction is connected to action, and when the predictive model is used in the learning lifecycle. Separating these questions clarifies why methods that appear similar can serve different roles, and why methods with different architectures can address the same functional bottleneck. Across latent dynamics, video prediction, flow, geometry, physics-informed modeling, learned environments, and VLA systems, the central criterion is not prediction quality in isolation. A world model is useful when its predictions remain actionable, physically credible, calibrated, and aligned with the policy, planner, simulator, or verifier that consumes them.

The main challenge ahead is evaluation under intervention. Current benchmarks still underspecify contact, force, deformation, long-horizon physical consistency, and recovery from failure. Metrics remain fragmented across visual fidelity, task success, sample efficiency, policy ranking, and simulator reliability. As world models become active substrates for post-training search and verification, the field needs benchmarks that directly test hallucination, action alignment, closed-loop policy exploitation, and sim-to-real trustworthiness. Progress will depend less on making imagined futures look plausible and more on making them reliable enough to improve real robot learning and decision making.

Several practical directions follow from this view. Datasets should include not only successful demonstrations, but also failures, recoveries, contact-rich transitions, and non-visual signals such as force, tactile, audio, event, and proprioceptive streams. Model evaluations should report both downstream utility and predictive fidelity, and should include ablations that separate representation learning from rollout-based reasoning. Learned simulators should be judged by policy-rank correlation, robustness to rollout depth, and resistance to exploitation, rather than solely by visual realism. Finally, future world models should expose enough structure for verification, whether through geometry, physical constraints, uncertainty estimates, or independent feasibility checks. These requirements point toward hybrid systems that combine scalable predictive learning with explicit mechanisms for grounding, calibration, and closed-loop trust.

\section{Acknowledgements}
This work was supported by the National Natural Science Foundation of China (NSFC) under Grant No. 62403211, and in part by Youth S\&T Talent Support Programme of Guangdong Provincial Association for Science and Technology (GDSTA) under Grant No. SKXRC2025092.

\bibliographystyle{IEEEtran}
\bibliography{survey}

\end{document}